\documentclass[final]{clv2025}

\jvol{vv}
\jnum{nn}
\jyear{2025}

\renewcommand{\footmark}{}
\jname{Zhang and Sheng}
\jinfo{Calibrated Tests on the Voynich Manuscript}

\usepackage{amsmath}
\usepackage{booktabs}
\usepackage{latexsym}
\usepackage[T1]{fontenc}
\usepackage[utf8]{inputenc}
\usepackage{microtype}
\usepackage{inconsolata}
\usepackage{graphicx}

\usepackage{amsmath}
\usepackage{amssymb}
\usepackage{booktabs}
\usepackage{array}
\usepackage{xcolor}
\runningtitle{Calibrated Tests on the Voynich Manuscript}
\runningauthor{Zhang and Sheng}

\begin{document}
	
	\title{Can Statistics Tell Whether an Undeciphered Text Is Language? Calibrated Tests on the Voynich Manuscript}
	
	\author{Xin Zhang\thanks{Corresponding author.}$^{,1}$ and Victor S. Sheng$^{1}$}
	
	\affilblock{
		\affil{Texas Tech University, Lubbock, TX, USA\\\quad \email{\{xin.zhang, victor.sheng\}@ttu.edu}}
	}
	
	\maketitle
	
	\begin{abstract}
		Statistics are often the only evidence of whether a text in an unread script records a language, but meaningless text shares many of them. A test means little until it is run on texts whose answer is known. We build such tests and apply them to the Voynich manuscript, whose text may be meaningless copying or a language hidden by a cipher. We write both accounts as programs, fit them with the same search budget and check them on held-out statistics. The copying process of Timm and Schinner predicts these statistics better than any cipher with one key, but ciphers with two keys, or that encode only the word classes of a real text, do equally well. A trace of language that needs no decipherment says more. In 28 texts in 16 natural languages the same word pairs keep sharing a line, and the manuscript shows this at half the strength of the weakest. Word codes fall to this level only with heavy miscopying or a plaintext whose word pairs recur weakly, such as abbreviated Latin. Solvers that read enciphered controls find nothing readable in the manuscript under letter ciphers, word codes or codes with one word per letter or syllable. One cipher remains, a nomenclator with two keys, which writes frequent words whole and spells the rest, and whose scribe miscopies more than half of the words. Our tests exclude it in all languages but Arabic and Hebrew. Planted messages show that a short message could still hide.
	\end{abstract}
	
	\section{Introduction}
	
	Some texts are written in scripts that no one can read. Before anyone tries to decipher such a text, it helps to know whether it records a language at all, and statistics of the text are often the only evidence. Whether such statistics can show this has been disputed in \textit{Computational Linguistics}. \citet{sproat2010ancient} argued that entropy measures, offered to show that the Indus script and the Pictish symbols write a language, cannot tell writing from other systems of symbols, and \citet{rao2010entropy} replied that such measures make a language more likely without proving it. \citet{tanakaishii2015constancy} found that a measure of how often units repeat places the Voynich manuscript and the Rongorongo script among natural languages, and noted that this adds only a little evidence. A statistic says little about an unknown text until it has been measured on texts whose answer is known, both texts with language and texts without. Decipherment has the same problem. Solvers built on language models have read many ciphers \citep{ravi2011zodiac,nuhn2013beam,bergkirkpatrick2013decipherment}, but a solver that finds nothing in an unknown text tells us something only if it reads texts made by the cipher it assumes. If statistics and solvers are first tested on such texts, what can they tell us about whether an unread text records a language?
	
	The Voynich manuscript (Beinecke MS 408) is a book of about 240 pages that radiocarbon dating places in the early fifteenth century \citep{bowern2021linguistics}. Its drawings of plants, stars and bathing figures surround some 38,000 words in a script found nowhere else. The text looks like writing. Its words are separated by spaces, its vocabulary has a long tail of rare words, and its glyphs keep to fixed places inside words. But it behaves unlike any known language. The same word often appears two or three times in a row, many words differ from their neighbours by a single glyph, and the frequent words are short and resemble each other \citep{reddy2011voynich,bowern2021linguistics}.
	
	Two accounts try to explain this text. The first holds that it means nothing. A scribe wrote each new word by copying an earlier word, often one close by, and changing a glyph or two \citep{timm2020generating}. Related proposals generate meaningless text with tables and grilles \citep{rugg2004hoax}, and volunteers asked to write gibberish produce text whose statistics resemble those of the manuscript \citep{gaskell2022gibberish}. The second account holds that the text is a real language hidden by a cipher. The Naibbe cipher \citep{greshko2025naibbe} shows that a scribe with the tools of the period could turn Latin or Italian into text that looks much like the manuscript.
	
	For a long time each account was defended by showing that its own procedure reproduces statistics chosen for the purpose. Recent work has started to test the accounts against controls. \citet{schwartz2026voynich} scores readings on pages held back from the fit, \citet{rozanova2026glyph} run ciphers and generators through the same tests, and \citet{parisel2026layered} tests two kinds of generator against several criteria at once. Three questions remain open. Do the two accounts fit equally well under the same conditions, and which of them predicts what it was not tuned to? Does the text carry any trace of language that we can see without deciphering it? And could a short message hide inside a text that is otherwise copied?

	Most answers point against a cipher of an ordinary text. The copying process predicts the held-out statistics better than any cipher with one key for the whole book, but ciphers with two keys, or ciphers that encode only the word classes of a real text, do equally well. The four published measures cannot decide either. They are the two entropy measures offered for the Indus script and the Pictish symbols \citep{rao2009entropic,lee2010pictish}, the measure of how often units repeat \citep{tanakaishii2015constancy}, and the share of repeated signs that directly follow themselves \citep{sproat2014statistical}. Copying programs reach the manuscript on all four, as texts with language do. These measures respond to how regular a text is, and copying makes a text regular too. The trace of language says more, since copying has no reason to produce it. The manuscript shows it at half the strength of the weakest of our 28 texts, and word codes of most texts keep it far above the manuscript. Solvers that decipher their controls find no readable text in the manuscript. A word code of abbreviated Latin has word pairs that recur about as weakly as those of the manuscript. A solver with a model of 5 million words of such Latin still finds clear word order in enciphered controls, and finds as little as in the manuscript only when 40 percent of their words are miscopied. One cipher escapes these tests, a nomenclator that writes frequent words whole and spells the others letter by letter. With two keys and a scribe who miscopies more than half of the words, it matches the word pairs of the manuscript and the held-out statistics. A solver built for this cipher excludes the two best-fitting settings in 14 of the 16 languages, and the third writes one word far more often than the manuscript writes any word. The cipher remains open only in Arabic and Hebrew. Our tests catch every planted message of 500 letters among the first words of paragraphs, but only if each letter has a fixed word and they are adjacent. Nothing in the manuscript points to such a message. The strong link between the last glyph of a word and the first glyph of the next carries most of the structure that our letter solver finds in the text. It reflects glyphs bound to their neighbours twice as tightly as letters in the languages we tested, and no program reproduces both this link and the manuscript's spaces. It does not come from where words stand in the line, and unlike the link in every plaintext it ends at the line break.
	
	We make four contributions, each of which also applies to other undeciphered texts (Section~\ref{sec:discussion}).
	
	(1) A comparison of explanations under equal conditions (Section~\ref{sec:programs}). We write each account as a program that fills the pages and lines of the manuscript word by word. Copying programs and ciphers draw new words from the same models of Voynich words and get the same search budget. Every cipher has a scribe who copies as the copying programs do, so local repetition favours neither side. Each program is judged on held-out statistics, which no program was fitted to.
	
	(2) A trace of language that needs no decipherment (Section~\ref{sec:language}). In real texts, two words that share one line tend to share other lines as well, and a code that replaces each word keeps this pattern. We calibrate this trace on 28 texts in 16 languages and on ciphers of these texts whose words are miscopied at known rates. With the texts of the copying programs, the same controls also test four published measures of language (Section~\ref{sec:published}).
	
	(3) Decipherment calibrated on controls (Section~\ref{sec:decipher}). Every solver must first read texts enciphered in the way it assumes, and a test counts as undecided when it cannot. Texts that keep the code but scramble the plaintext show how much a solver finds without any language. We apply this to letter ciphers, word codes, codes that mix whole words with spelled ones, and 28 codes with one letter, letter pair or syllable per word.
	
	(4) A bound on what could still hide (Section~\ref{sec:short}). We plant messages of known length in the text and record from what length a fixed search finds them all, so a search that finds nothing still limits what the text could contain. We also test other ways the text could carry meaning, from filler words to a numbering of the zodiac figures (Section~\ref{sec:other}).

	\section{Related Work}
	\label{sec:related}

	\paragraph{Statistical evidence of language}
Statistics have long been offered as evidence that an unread script writes a language. \citet{rao2009entropic} found that the conditional entropy of neighbouring signs places the Indus script near written languages such as Sumerian and Old Tamil and away from DNA and proteins. \citet{lee2010pictish} classified the Pictish symbols as writing with a decision tree built on entropy measures. \citet{sproat2010ancient,sproat2010reply} argued that such measures do not separate writing from other symbol systems and that texts made at random can match them, and both groups defended their measures \citep{rao2010entropy,lee2010response}. Comparing written languages with many nonlinguistic symbol systems, \citet{sproat2014statistical} found that none of the published measures separates the two, while a measure of repetition and the measure of \citet{lee2010pictish} with a new threshold do. Both of these call the Indus and Pictish symbols nonlinguistic. \citet{raghavendra2026nonspecificity} built an emblem system that carries meaning but writes no speech. On the six published tests whose decision rule and result for the Indus script could be reproduced, it receives the same verdict as the Indus script, and the author concludes that such measures detect structure but cannot by themselves show that a script records speech. Measures of this kind have also been applied to the Voynich manuscript. Its words follow Zipf's law and cluster by section the way keywords do \citep{montemurro2013keywords}, and network, entropy and spectral measures place it nearer to natural language than to random text \citep{landini2001spectral,amancio2013probing,tanakaishii2015constancy}. Topic models follow its scribes and subjects \citep{sterneck2021topic}, and latent semantic analysis finds a coherence like that of known texts \citep{layfield2026lsa}. The information carried by the parts of a word also varies with the word's probability as in natural languages \citep{layfield2020word}. Most of these studies set the manuscript beside natural texts or beside shuffled and random text. \citet{gaskell2022gibberish} also compared it with gibberish written by volunteers and found it closer to gibberish than to meaningful text. A statistic points to language only if meaningless text made by a plausible process does not reach it as well. We calibrate our tests on such texts and on texts with language. Calibrated in this way, the conditional entropy, the entropy classifier, the repetition measure and Yule's K \citep{tanakaishii2015constancy} cannot tell copying from language in the manuscript (Section~\ref{sec:published}).

\paragraph{Decipherment}
A solver for a cipher searches for the key under which a language model scores the plaintext highest. It has solved homophonic ciphers \citep{ravi2011zodiac,nuhn2013beam,bergkirkpatrick2013decipherment} and historical ciphers such as the Copiale cipher \citep{knight2011copiale}, and it gains from neural language models \citep{kambhatla2018decipherment,aldarrab2021seq2seq}. Related models decipher a lost language through a known relative, such as Ugaritic through Hebrew and Linear B through Greek \citep{snyder2010lost,luo2019neural}, even when the script does not mark word breaks \citep{luo2021deciphering}. \citet{hauer2016anagrammed} identified the language of synthetic ciphers in 380 languages and proposed Hebrew for the manuscript. All of these methods are judged on texts whose plaintext is known, and their accuracy falls as the ciphertext gets shorter \citep{ravi2008attacking}. A solver that finds nothing in an unknown text therefore says something only about ciphers that it has been shown to read at that length. We hold our solvers to this standard, with controls made by the cipher in question, about as long as the manuscript and miscopied at the fitted rate.

\paragraph{Word co-occurrence}
Collocation measures test whether two words co-occur more often than chance \citep{church1990association,dunning1993accurate}. A word that has appeared once in a document is also far more likely than chance to appear again, above all a content word \citep{church2000empirical}. Copying produces such repetition as well, so a test for language must not take repeated or nearly repeated words as evidence. Our test leaves out pairs of similar words and asks whether the pairs that share lines in one half of the book also share lines in the other, compared with words shuffled only within pages (Section~\ref{sec:language}).

\paragraph{Testing models that generate text}
Our programs have no tractable likelihood. We fit them by comparing statistics of simulated and observed text, as in approximate Bayesian computation \citep{beaumont2002abc,toni2009abc}, and judge the fits with predictive checks \citep{gelman1996posterior} on held-out statistics. Language models are judged in a similar way, by whether the texts they generate follow the statistical laws and tendencies of natural language \citep{takahashi2019evaluating,meister2021language}. We plant messages of known length and count how many a fixed search recovers, as in the injection--recovery tests of planet searches \citep{christiansen2013kepler}.

\paragraph{Statistics of the manuscript}
\citet{currier1976findings} found two dialects, now called Currier A and B, with different glyph and word preferences, and noted that the line behaves like a unit of the text. \citet{davis2020scribes} distinguished five scribal hands. \citet{smith2019glyph} found that some glyphs on the two sides of a word break meet far more often and others far less often than chance predicts. \citet{feaster2022rightward} showed that glyphs prefer places in the line and in the paragraph and argued that this explains some of these pairs. The glyph sequences are unusually predictable \citep{lindemann2020entropy,bowern2021linguistics}, slightly more so when read from right to left \citep{parisel2025directionality}, and some glyphs may be ligatures of others, which would make the alphabet much smaller \citep{matlach2022symbol}. Word lengths depend on the drawings and on line position \citep{steckley2024subtle}. These studies describe the text. They do not fit a process that produces it.
	\begin{table*}[t]
	\centering
	\footnotesize
	\setlength{\tabcolsep}{3pt}
	\newcommand{\hd}[2]{\begin{tabular}[b]{@{}c@{}}#1\\#2\end{tabular}}
		\caption{What studies of the two accounts test. Meaningless text means human gibberish or a generator, ciphers means enciphered real text, held-out test means that a program is fitted on some data and judged on other data, decipherment on controls means that a solver is first checked on enciphered texts with known plaintext, noise means that the enciphered controls contain scribal or transcription errors, and short messages means that messages of known length are planted and searched for.}
	\label{tab:related}
	\begin{tabular}{@{}lcccccc@{}}
		\toprule
		Study & \hd{Meaningless}{text} & \hd{}{Ciphers} & \hd{Held-out}{test} & \hd{Decipherment}{on controls} & \hd{Noise in}{the cipher} & \hd{Short}{messages} \\
		\midrule
		\citet{timm2020generating} & \checkmark & & & & & \\
		\citet{hauer2016anagrammed} & & \checkmark & & \checkmark & & \\
		\citet{gaskell2022gibberish} & \checkmark & & & & & \\
		\citet{bowern2022enciphered} & \checkmark & \checkmark & & & & \\
		\citet{greshko2025naibbe} & & \checkmark & & & & \\
		\citet{parisel2026layered} & \checkmark & & & & & \\
		\citet{rozanova2026glyph} & \checkmark & \checkmark & & \checkmark & & \\
		\citet{schwartz2026voynich} & \checkmark & \checkmark & \checkmark & \checkmark & \checkmark & \\
		This work & \checkmark & \checkmark & \checkmark & \checkmark & \checkmark & \checkmark \\
		\bottomrule
	\end{tabular}

\end{table*}

\paragraph{Two accounts of the manuscript}
\citet{rugg2004hoax} generated words from a table of word parts with a grille, and \citet{rugg2017hoaxing} argued that the statistics of the text follow from such a method. \citet{schinner2007hoax} argued from the statistics of the text for a mechanical origin. \citet{timm2020generating} proposed the copying process (self-citation) and released a generator \citep{timm2019generator} whose output shares many statistics with the manuscript. They later reviewed competing hypotheses \citep{timm2024discussion}, and \citet{timm2026dynamic} argues that the text changes gradually as it grows, which defeats any analysis that assumes a fixed system. \citet{boxer2022fingerprinting} showed that the manuscript does not belong to the same family as a known gibberish, the angelic texts of Dee and Kelley. \citet{izumo2026shelf} built a physical machine that reproduces word lengths and low-order glyph transitions, and \citet{turenne2026pastiche} models the text with position-dependent grammars and reads it as a pastiche of language and of medieval medical books. For the cipher account, \citet{bowern2022enciphered} applied 22 manipulations to Latin and found that some encodings with bigraphs and nulls bring it into the range of the manuscript, though no single one matches every metric. \citet{greshko2025naibbe} designed the Naibbe cipher, which cuts the plaintext into letters and letter pairs and writes each as one of several Voynich-like words chosen with cards and dice. Its ciphertext matches many statistics of Currier B but not the vocabulary of the manuscript or its line effects.

\paragraph{Calibrated tests of the two accounts}
Three recent studies come closest to ours, and Table~\ref{tab:related} sets them beside earlier work. \citet{parisel2026layered} generated text from pools of word prefixes and suffixes, and with grilles in the manner of \citet{rugg2004hoax}, and found that no setting of either generator reproduces four statistics at once. One is a link across word boundaries of 0.230 bits in the manuscript against 0.054 to 0.173 in four natural languages. \citet{schwartz2026voynich} fits every model on half of each scribe's pages and scores it on the other half. He tests 130 reading hypotheses in 18 languages and sets the acceptance rule on real texts in six languages, disguised by 16 mechanisms at four levels of transcription noise. The generator of Timm and Schinner fails all 7,030 trials, no reading survives except where the test lacks power, and a keyless procedure of word reuse and coinage violates only two or three of 22 statistics. \citet{rozanova2026glyph} apply the same tests to prose in five languages, to simple, verbose, homophonic and Naibbe ciphers and the generator of Timm and Schinner. Their substitution attack merges frequent glyph sequences into units by byte-pair encoding, sorts the units into 23 classes and maps each class to a letter. It recovers 74 to 93 percent of the key of synthetic Latin ciphers and finds nothing in the manuscript. All three studies report a link between glyphs across word boundaries that no generator reproduces. We find the same link, show that it carries most of the structure that a solver finds in the text, and trace its strength to glyphs bound twice as tightly as letters.

Our study differs from these in four ways. We fit copying programs and ciphers with the same search budget, let them draw new words from the same models of Voynich words, and give our ciphers the same copying scribe as the copying programs, so that local repetition favours neither side. We measure a trace of language that needs no decipherment and calibrate it on 28 texts in 16 languages and on ciphers miscopied at several rates. We check every solver on texts that keep the code but scramble the plaintext, which shows how much a solver finds without any language, and we report a test as undecided when its controls cannot be read. Finally, none of these studies plants short messages, so none can say how long a hidden message would have to be before it shows.
	
	\section{Data}
	\label{sec:data}
	
	\paragraph{Manuscript}
	We use two transliterations in the EVA alphabet \citep{zandbergen2025transliteration}. We fit every program to ZL (version 3b, by Zandbergen and Landini) and use IT (version 2a, extracted from the Landini--Stolfi interlinear file) as a check. We keep running text and the text written along circles and radii, read uncertain spaces as word breaks and split words into glyphs by the EVA conventions, under which ch, sh, cth, ckh, cph and cfh count as single glyphs. This gives 37,554 tokens for ZL and 36,736 for IT. The transliteration gives the Currier dialect, scribe, section and quire of every page. We also use the scans that the Beinecke Library publishes (Table~\ref{tab:sources}) to measure the plant drawings and the gaps between neighbouring glyphs. On these scans three in four of the spaces that ZL marks as uncertain are narrower than a word space (Section~\ref{sec:link}).
	
	\paragraph{Texts in sixteen languages}
	Our main corpus covers Latin, Italian, German, Middle English and Middle French with texts written between 1300 and 1500. The Italian and Middle French texts are machine transcriptions of manuscripts from CoMMA \citep{clerice2026comma}. For the letter solver's 16 language models (Section~\ref{sec:reading}), both languages come from edited texts and early prints of the same period instead (Appendix~\ref{app:corpora}). German comes from Baumbank.UP \citep{demske2019baumbank}, and Middle English from CME, PCEEC and the Helsinki Corpus \citep{mcsparran2018cme,taylor2006pceec,rissanen1991helsinki}. A further corpus gives Latin transcribed with its abbreviations as written \citep{clerice2023cremma,glaise2023htromance,parisbible}, and the Latin manuscripts of CoMMA give more such text for the larger language models of Section~\ref{sec:wordcodes}.
	Eleven further languages come from public collections, namely medieval Hebrew, classical Arabic and classical Persian written as consonants, Byzantine Greek, Old Church Slavonic with Old East Slavic, Old French, Old Occitan, Old Spanish, Galician-Portuguese, Old Norse and early modern Czech, the last from the Kralice Bible because we found no medieval Czech text. Greek and Slavonic are transliterated into Latin letters, and every text except abbreviated Latin is reduced to the letters a to z. Each language is split by document into training text for language models and held-out text for controls, so that no document contributes to both. Appendix~\ref{app:corpora} lists the sources and sizes. We count sixteen languages, with abbreviated Latin as a seventeenth variant in some tests. The ciphers of Section~\ref{sec:programs} choose their plaintext among the five languages of the main corpus and abbreviated Latin. The shuffled manuscripts of Section~\ref{sec:reading} and the medical texts of Section~\ref{sec:wordcodes} use only these five languages. The test of halves of words in Section~\ref{sec:other} leaves out German, which has too little text.
	
	\section{Two Accounts Under Equal Conditions}
	\label{sec:programs}
	
	\subsection{Programs}
	
	Every program writes into the layout of ZL. It fills the same pages and lines with the same number of words, so layout cannot separate the accounts. Table~\ref{tab:programs} lists the programs.
	
	\begin{table}[t]
		\centering
		\small
		\setlength{\tabcolsep}{2.5pt}
				\caption{The programs. Shapes come from the model W of Voynich words (with an optional second setting for Currier B in Cipher-W2, Cipher-S, Cipher-C and Cipher-M2), the Naibbe tables or a table of word parts. Changes are W edits, the rules of Timm and Schinner with or without five writing habits, or the scribe of the ciphers. The last column counts continuous parameters plus the choice of plaintext. The three programs at the bottom are fitted to a larger set of statistics (Section~\ref{sec:morecopying}). The letters name what each program rests on, W the word model, T Timm's rules, N the Naibbe tables, S syllables, C word classes, M a nomenclator, H habits and L a lexicon, and W2 a second key for Currier B.}
		\label{tab:programs}
		\begin{tabular}{@{}llllr@{}}
			\toprule
			Program & Kind & Shapes & Changes & Free \\
			\midrule
			Copy-W & copying & W & W edits & 10 \\
			Copy-T & copying & W & Timm's rules & 12 \\
			Copy-N & copying & Naibbe & W edits & 8\,+\,1 \\
			\midrule
			Cipher-W & word code & W & scribe & 15\,+\,1 \\
			Cipher-N & letter cipher & Naibbe & scribe & 9\,+\,1 \\
			Cipher-W2 & word code & W, twice & scribe & 18\,+\,1 \\
			Cipher-S & syllable code & W, twice & scribe & 18\,+\,1 \\
			Cipher-C & word classes & W, twice & scribe & 19\,+\,1 \\
			Cipher-M & nomenclator & W & scribe & 16\,+\,1 \\
			Cipher-M2 & nomenclator & W, twice & scribe & 19\,+\,1 \\
			Cipher-MN & nomenclator & W, Naibbe & scribe & 16\,+\,1 \\
			\midrule
			Copy-H & copying & W & Timm, habits & 17 \\
			Copy-L & lexicon & W & Timm, habits & 21 \\
			Grille & grilles & parts & Timm, habits & 19 \\
			\bottomrule
		\end{tabular}

	\end{table}
	
	\paragraph{Shared word shapes}
	Programs that need new words draw them from one model W of Voynich words, a glyph trigram model with Witten--Bell smoothing \citep{witten1991zero} trained once on the word types of ZL. We sample a pool of 400,000 words from W once. Each simulation reweights this pool with two parameters, a pull towards short words and an exponent that sharpens the preference for typical forms. A W edit changes, inserts or deletes one glyph. It first draws the kind of edit, then scores eight random candidates by their W probability relative to typical words of the same length and keeps the best. The Naibbe programs take their word shapes from the tables of \citet{greshko2025naibbe}.
	
	\paragraph{Copying programs}
	A copying program writes each word as a new word or as a copy of a word already written. The source of a copy is the word above, a random earlier word of the page, a word at a geometrically distributed distance back, or, on the first line of a page, a word of the previous page. After a copy the program may go on to copy the word that follows the source, which reproduces stretches of earlier text. A copy stays exact with some probability and is changed otherwise. Copy-W changes copies with W edits and draws new words from W. Copy-N draws new words by enciphering random letters and letter pairs of a language with the Naibbe tables, so they carry no meaning.
	
	Copy-T changes copies with the rules of \citet{timm2020generating}, which we port from their published generator \citep{timm2019generator}. These rules split a word into glyph groups such as qo, ee, iin and ol. A change replaces one to three groups by similar groups with Timm's probabilities, and the last group of a word has its own table. Other changes add or remove a prefix or gallows glyph (k, t, p or f), and others join the front of one word to the back of the next or split a long word. Timm's rules for the first word of a line and the first line of a paragraph apply, and a changed word may be followed by part of itself. Timm's generator decides with hand-written rules which glyph may follow which. We use its statistical option in the simplest form, which accepts a pair of glyphs if the pair occurs inside a word of ZL. The shares of the three kinds of change are free parameters with Timm's defaults inside their ranges, and new words come from W.
	
	\paragraph{Cipher programs}
	A cipher program enciphers a real plaintext unit by unit. Cipher-W uses a word codebook. A plaintext word $w$ with frequency $f(w)$ receives $1+\lfloor h\,f(w)^{\alpha}\rfloor$ homophones drawn from W. Frequent words receive the most typical forms with strength $\gamma$, and each further homophone of a word becomes a one-glyph variant of an earlier homophone with a fitted probability. Cipher-N uses the Naibbe cipher, which cuts the plaintext into letters and letter pairs and writes each unit as one of its homophones from Greshko's tables. We added three ciphers after the first results. Cipher-W2 gives the pages of Currier B a second key with its own word-shape parameters, and a fitted share of units keeps the same homophones in both keys. Cipher-S writes one word for each plaintext syllable. Cipher-C keeps only the word classes of the plaintext. It clusters frequent words by their contexts into a fitted number of classes, puts rare words into three classes by length, and writes the sequence of classes with homophones from W. The fit may also give Cipher-S and Cipher-C a second key. Last, we added three nomenclators, a kind of code common in the period. A nomenclator writes the most frequent words of its plaintext as whole code words and spells all other words letter by letter, one written word per letter. The fit chooses the number of code words between 5 and 2,000. Cipher-M takes the code words and the homophones of the letters from Cipher-W, Cipher-M2 adds the second key of Cipher-W2, and Cipher-MN spells the other words with the Naibbe tables. With every word coded, Cipher-M and Cipher-M2 would write the texts of Cipher-W and Cipher-W2, and without code words Cipher-MN would write those of Cipher-N.
	
	All cipher programs add a scribe who chooses among the homophones of each unit. With a probability we call the copy bias, he picks a homophone within one glyph of a word he looks at. He looks where the copying programs copy from, the word after the one his last choice came from, a word of the previous page, the word above and earlier words of the page. The scribe also writes null words, miscopies words with a W edit and now and then writes a word twice. All other words still decode to exactly one plaintext unit, so the text keeps its meaning. This scribe gives the ciphers the same means of local repetition that the copying programs have.
	
	\subsection{Fitting and checking}
	\label{sec:fitting}
	
	\paragraph{Statistics}
	We describe a text by 22 fitted statistics in four groups (Appendix~\ref{app:stats}). Eleven measure copying. They cover how often a word repeats or nearly repeats a word one to twenty positions earlier, how often it resembles the word above it or a word far away on its page, how specific the vocabulary of a page is, and how word types one glyph apart form a network. Seven describe the vocabulary and word lengths, one measures word order as the mutual information between adjacent words, and three describe glyph sequences. Statistics that depend on position are measured against the same text with its words shuffled inside each page. Fifteen held-out statistics never enter the fit. Eleven describe the growth of the vocabulary, its frequency spectrum, the shares of short and long words, how far back a word last appeared, near repeats among the ten previous words, repeats within a line, reused word pairs, the entropy of a word given the previous one and the overlap of neighbouring pages. Four test the two accounts directly. The first asks whether word types one glyph apart occur in the same contexts, using positive pointwise mutual information \citep{church1990association}. The second is the slope of the vocabulary overlap of two pages against their distance, which measures the drift that \citet{timm2026dynamic} describes. The last two are Jensen--Shannon divergences \citep{lin1991divergence} between Currier A and B pages, one for word frequencies and one for glyph pairs inside words. Four layout statistics, such as the glyph preference of line-initial words, enter only the fits of Sections~\ref{sec:morecopying} and~\ref{sec:link}. All statistics were fixed before the fits reported here.
	
	\paragraph{Search}
	The distance between a simulated text $x$ and the manuscript is
	\begin{equation}
		D(x)=\Bigl(\frac{1}{22}\sum_{k=1}^{22}\Bigl(\frac{s_k(x)-s_k(\mathrm{ZL})}{\sigma_k}\Bigr)^{2}\Bigr)^{1/2},
	\end{equation}
	where $s_k$ is the $k$-th fitted statistic and $\sigma_k$ its robust spread over the first-round simulations of the first five programs. For each program we draw 1,000 parameter sets from the prior and then run six rounds of 500 proposals around the 100 simulations closest to the manuscript so far. Every program thus receives 4,000 simulations. The 25 best distinct parameter sets are simulated four more times with new seeds, and these 100 texts form the predictive sample of the program. We run the whole search twice with independent seeds (priors in Appendix~\ref{app:priors}).
	
	\paragraph{Checks}
	For every statistic we record whether the manuscript lies inside the central 95\% of the predictive sample and its $z$-score with respect to the sample mean and standard deviation. When it does, we say that the program covers the statistic. For a set of statistics we compute the Mahalanobis distance of the manuscript from the predictive sample with a shrinkage covariance \citep{ledoit2004shrinkage} and rank it among the leave-one-out distances of the 100 sample texts. The rank gives a predictive p-value, which with 100 texts cannot fall below 0.01. A program passes a check when its p-value exceeds 0.05.

	\subsection{Results}
	\label{sec:results}
	
	\begin{table*}[t]
		\centering
		\scriptsize
		\setlength{\tabcolsep}{2.5pt}
				\caption{Fit of the programs to ZL, first and second search. Best is the smallest distance among the 4,000 simulations of a search, and median the median distance of its 100 predictive texts. Among each program's own texts, the Mahalanobis distance over the 15 held-out statistics has a 95th percentile of 4.2 to 5.2. The p-values are predictive p-values of at least 0.01, for the held-out statistics and for all 37 statistics. The last row fixes the number of word classes of Cipher-C at two.}
		\label{tab:results}
		\begin{tabular}{@{}lcccccccc@{}}
			\toprule
			& \multicolumn{2}{c}{Distance $D$} & \multicolumn{2}{c}{Inside predicted range} & \multicolumn{3}{c}{Held-out statistics} & All 37 \\
			\cmidrule(lr){2-3}\cmidrule(lr){4-5}\cmidrule(lr){6-8}\cmidrule(lr){9-9}
			Program & best & median & fitted (22) & held-out (15) & largest $|z|$ & Mahalanobis & p-value & p-value \\
			\midrule
			Copy-W & 0.69 / 0.65 & 0.75 / 0.74 & 13 / 18 & 10 / 11 & 14 / 15 & 27 / 27 & 0.01 / 0.01 & 0.01 / 0.01 \\
			Copy-T & 0.48 / 0.49 & 0.63 / 0.62 & 22 / 20 & 9 / 10 & 4.1 / 4.9 & 7.0 / 8.3 & 0.01 / 0.01 & 0.01 / 0.01 \\
			Copy-N & 0.47 / 0.49 & 0.53 / 0.55 & 12 / 14 & 2 / 3 & 107 / 77 & 87 / 79 & 0.01 / 0.01 & 0.01 / 0.01 \\
			\midrule
			Cipher-W & 0.50 / 0.52 & 0.62 / 0.61 & 13 / 13 & 8 / 4 & 101 / 147 & 78 / 87 & 0.01 / 0.01 & 0.01 / 0.01 \\
			Cipher-N & 0.36 / 0.37 & 0.41 / 0.42 & 12 / 14 & 6 / 4 & 219 / 188 & 90 / 90 & 0.01 / 0.01 & 0.01 / 0.01 \\
			Cipher-W2 & 0.52 / 0.56 & 0.70 / 0.69 & 18 / 14 & 13 / 13 & 5.2 / 4.0 & 9.9 / 8.2 & 0.01 / 0.01 & 0.01 / 0.01 \\
			Cipher-S & 0.60 / 0.56 & 0.68 / 0.70 & 18 / 19 & 10 / 11 & 8.2 / 4.8 & 12 / 8.3 & 0.01 / 0.01 & 0.01 / 0.01 \\
			Cipher-C & 0.55 / 0.61 & 0.70 / 0.72 & 20 / 20 & 15 / 15 & 1.2 / 1.2 & 3.6 / 3.0 & 0.32 / 0.59 & 0.01 / 0.02 \\
			Cipher-M & 0.65 / 0.64 & 0.74 / 0.76 & 20 / 21 & 11 / 11 & 55 / 34 & 59 / 38 & 0.01 / 0.01 & 0.01 / 0.01 \\
			Cipher-M2 & 0.54 / 0.58 & 0.71 / 0.74 & 20 / 21 & 15 / 15 & 1.6 / 1.6 & 4.2 / 3.5 & 0.17 / 0.41 & 0.01 / 0.01 \\
			Cipher-MN & 0.32 / 0.35 & 0.43 / 0.44 & 18 / 15 & 8 / 6 & 196 / 186 & 91 / 88 & 0.01 / 0.01 & 0.01 / 0.01 \\
			\midrule
			Cipher-C, two classes & 0.56 / 0.61 & 0.71 / 0.74 & 22 / 22 & 15 / 13 & 1.7 / 3.5 & 3.8 / 5.7 & 0.15 / 0.03 & 0.05 / 0.01 \\
			\bottomrule
		\end{tabular}

	\end{table*}
	
\begin{figure*}[t]
		\centering
		\includegraphics[width=\textwidth]{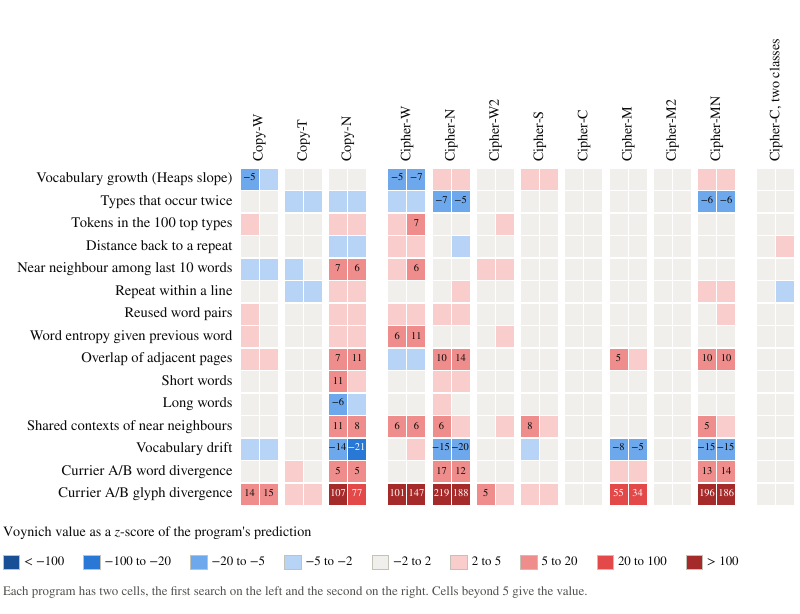}
		\caption{The manuscript (ZL) on the 15 held-out statistics, as a $z$-score against the predictive sample of each program in Table~\ref{tab:results}. A grey cell ($|z|\le 2$) means that the program predicts the manuscript on that statistic. Red means that the manuscript lies above the prediction and blue that it lies below, and darker colours mean larger misses. Appendix~\ref{app:z} gives all values.}
		\label{fig:heldout}
	\end{figure*}

	Table~\ref{tab:results} summarises the fits, and Figure~\ref{fig:heldout} shows the held-out statistics for every program.
	
	\paragraph{Copying against one-key ciphers}
	Cipher-W lies closer to the manuscript than Copy-W, and Cipher-N closer than Copy-N. Copy-W kept the narrower priors of the first fits, and its exponent of W ends near their upper bound of 12. Timm's rules change this comparison. Copy-T fits as well as Cipher-W, with a best distance of 0.48 and 0.49 against 0.50 and 0.52, and in about half of all pairs of predictive texts the cipher text is the closer one. Copy-T covers all 22 fitted statistics in the first search and 20 in the second. The held-out statistics separate the two sharply. Copy-T misses none of them by more than five standard deviations, while Cipher-W misses one by more than a hundred and Cipher-N by about two hundred.
	
	The statistic that separates them is the dialect difference in glyph use. Between Currier A and B pages the manuscript shows a divergence of 0.093 bits in its glyph pairs. Copy-T predicts about a third of it, and its 95\% range reaches 0.087 and 0.077. Cipher-W and Cipher-N predict 0.003 to 0.005. A program that draws its words from one fixed key writes similar glyph statistics on every page, whereas a scribe who builds almost every word from words he has just written lets his glyph habits wander through the book. Cipher-W also misses the shared contexts of similar words in both searches, and Cipher-N misses the drift of the vocabulary as well (Appendix~\ref{app:z}).
	
	\paragraph{Ciphers that close the gap}
	A second key can explain the dialect difference. Cipher-W2 gives the pages of Currier B their own word shapes. It predicts the glyph difference as well as Copy-T does (0.028 and 0.032), covers 13 of the 15 held-out statistics in both searches, and its Mahalanobis distance of 9.9 and 8.2 comes close to the 7.0 and 8.3 of Copy-T. Cipher-C does better than any other program, although its two best fits hardly use their second key. They give nearly every class the same homophones in both dialects. It covers all 15 held-out statistics in both searches, misses none by more than 1.2 standard deviations, and passes the held-out check with predictive p-values of 0.32 and 0.59. Its texts vary widely, from 0.003 to 0.34 bits in the glyph difference, so the manuscript falls inside its ranges easily. In the best simulations of Cipher-C the scribe miscopies about half of all words, and the best text of the first search uses only two word classes. When we fix the number of classes at two, so that counting the three classes of rare words the plaintext can carry at most 2.32 bits per word, the program still passes in one search and fails narrowly in the other (Table~\ref{tab:results}, last row). With 128 or 1,000 classes it passes in both searches, and with 16 classes in neither. The held-out statistics cannot tell how much of a plaintext such a text carries.
	
	The nomenclators show the same split. With one key, Cipher-M misses the glyph difference by 34 to 55 standard deviations and Cipher-MN by about 190. With two keys, Cipher-M2 comes within 1.6 standard deviations of all 15 held-out statistics in both searches and passes the held-out check with predictive p-values of 0.17 and 0.41. Its two best fits differ widely. The first codes only nine words and spells the rest with about 28 homophones per letter, so it is close to a letter cipher. The second codes 1,999 words of Latin, near the top of its range, and code words make up more than half of what it writes. The scribes of the two fits miscopy 37 and 56 percent of all words.
	
	No program passes the joint check over all 37 statistics. The best distances belong to Cipher-MN, which codes only five words, and to Cipher-N, but their texts vary so little that the manuscript falls outside their range on 4 to 10 fitted statistics. Distance and coverage can disagree, so we report both. Fits to IT rank the programs in the same order, and Cipher-C and Cipher-M2 pass the held-out check for IT as well ($p = 0.24$ / 0.51 and 0.10 / 0.21).
	
	\paragraph{The fitted scribes}
	The fitted parameters describe two kinds of scribe (Appendix~\ref{app:priors}). In the best simulations of Copy-T the median share of new words is about one percent, exact copies make up two percent and one percent of all copies in the two searches, and a copy continues with the next word of its source in nine cases out of ten. This scribe copies whole stretches of his page and changes nearly every word he copies. Cipher-W encodes the abbreviated Latin in nearly all of its best simulations and gives most plaintext words a single form, while Cipher-C and Cipher-M2 miscopy about half of their words. The ciphers that fit best are those whose plaintext matters least.
	
	\subsection{Three more copying programs}
	\label{sec:morecopying}
	
	Copy-T leaves three problems. Its texts link the last glyph of a word to the first glyph of the next far more weakly than the manuscript (mutual information 0.078 against 0.204 bits), its line-final words lack the glyph preferences of the manuscript, and a letter solver finds a third as much word order in its texts as in the manuscript (Section~\ref{sec:reading}). We built three programs to repair these failures and fitted them to 38 or 40 statistics, the 22 above plus the layout statistics, six measures of the coupling between neighbouring words, six shares of copying choices and, for Copy-L and Grille, two divergences of word use, between sections and between Currier A and B within the herbal section (Appendix~\ref{app:stats}). Their search scales each statistic by its robust spread over the first-round simulations of the same search.
	
	Copy-H adds five writing habits to Copy-T. It writes eight candidates for every word and picks one by weight. The weights favour the glyph pairs the manuscript prefers across word boundaries and at the start and end of lines, and favour or avoid repeating the first or last glyph of the previous word. The strengths of the habits are fitted, and their tables come from the manuscript itself. Copy-L replaces the new words of Copy-H with words from one lexicon for the whole book, weighted by dialect and section, and gives the labels their own lexicon. Grille reads words from a table of word parts through moving grilles in the manner of \citet{rugg2004hoax}, passes them through Timm's rules for the starts of lines and paragraphs, and keeps the five habits of Copy-H. In its search each scaled difference is capped at 5, so a statistic it cannot reach does not steer the search.
	
	All three fail, and more clearly than Copy-T. The manuscript lies inside the predicted range on only 4 to 10 of their 38 or 40 fitted statistics and on 0 to 3 of the 15 held-out statistics, with every joint p-value at 0.01. Copy-H still links neighbouring words only half as strongly as the manuscript (0.107 against 0.204 bits), makes pages too specific in their vocabulary and copies too often without change. Copy-L comes close to the link in one search but not in the other, and it keeps the other failures. Grille overshoots the link by a factor of almost three and repeats words a few positions back far too often. Copy-H fails in two further ways. Its pages, grouped by the five scribes of \citet{davis2020scribes}, differ more in vocabulary than the real scribes do in all ten pairs (for scribes 2 and 3, a divergence of 0.27 in the manuscript against 0.53 to 0.74 in the generated texts), and its labels repeat the text of their own page twice as often as real labels do (Section~\ref{sec:other}).
	
	\subsection{The link between neighbouring words}
	\label{sec:link}
	
	No program so far reproduces the link between the last glyph of a word and the first glyph of the next, which carries most of the structure that the letter solver of Section~\ref{sec:reading} finds. We tested three explanations under equal conditions. HABIT adds two habits to those of Copy-H. Its scribe favours the glyph pairs across word boundaries that he has written recently, a preference that grows from his own text and needs no table from the manuscript. He may also rewrite the first glyph of a word to suit the end of the previous one, with a table taken from the manuscript. SPACE writes a text of Copy-T, removes its spaces and puts them back by a local rule that uses the manuscript's odds of a space between two glyphs, its word lengths and a pull towards the old spaces, with the number of words in each line kept. STRING lets the scribe copy strings of glyphs instead of words. He writes each line as one string, piece by piece. A piece is either a new word from W or a copy of earlier glyphs that starts at a source chosen as in Copy-T. A copied piece may go on past a space of the earlier text into the next word, and it stays exact or is changed by Timm's rules. After each glyph he leaves a space or not. The chance of a space depends on how long the word is so far, on whether a new piece begins, on the glyphs next to the gap and on whether the word so far is ol, or, al or ar. It sees these glyphs only as four classes defined by their strokes. Unlike SPACE, STRING takes no table from the manuscript for where spaces fall or which glyphs meet across them. Its only table from the manuscript is the one Copy-H uses for the last glyph of a line. A first version without that table fitted far worse (best distance 3.3 against 0.92) and failed every check. We fitted each program twice to the 38 statistics of Section~\ref{sec:morecopying} and twice to the 32 that remain when the six measures of coupling are held out. The search is that of Section~\ref{sec:fitting}. For all three programs it scales each statistic by its robust spread over the first-round simulations of HABIT and SPACE, and the checks standardise each statistic before the Mahalanobis distance. Six measures of the spaces enter no fit. By the rules we fixed before the run, an explanation is rejected when both searches fail on the fitted link or both fail on the spaces.
	
		\begin{table*}[t]
		\centering
		\footnotesize
		\setlength{\tabcolsep}{3pt}
				\caption{The link between neighbouring words and the spaces. Link across the space is the mutual information (bits) between the last glyph of a word and the first glyph of the next inside lines, and ends of neighbours the same for the last glyphs of neighbouring words. Space from two glyphs is the share of the uncertainty about a space that the two glyphs around it remove, beyond the two glyphs is what a space says about the glyph that follows them (bits, beyond chance), and words also split is the share of tokens that also stand split into two neighbouring words, beyond chance. The plaintexts give their range, and the programs the central 95\% of their texts, both searches pooled. HABIT, SPACE and STRING were fitted without the six measures of coupling, Copy-H, Copy-L and Grille give 100 texts of each of their two best fits, no table means the same parameter sets and seeds with the manuscript's tables for the two glyphs around a space switched off, no crossing means the same for STRING with copied pieces that never cross a space, and no glyph rules means the same for STRING with the weights of its glyph classes and short words set to zero. All statistics means HABIT fitted to all 38 statistics, the six measures of coupling among them. Kept at a line break is the link across a line break inside a paragraph divided by the link inside lines, both beyond what the places of the words give (Section~\ref{sec:link}), in the running text and for the texts with a link inside lines. Bold marks ranges that exclude ZL.}
		\label{tab:link}
		\begin{tabular}{@{}lcccccc@{}}
			\toprule
					& Link across & Ends of & Space from & Beyond the & Words also & Kept at a \\
		& the space & neighbours & two glyphs & two glyphs & split & line break \\
		Texts & (bits) & (bits) & (share) & (bits) & (share) & (ratio) \\
			\midrule
			ZL & 0.204 & 0.027 & 0.73 & 0.022 & 0.186 & $-$0.003 \\
			IT & 0.187 & 0.026 & 0.74 & 0.023 & 0.101 & $-$0.013 \\
			\midrule
			29 plaintexts & \textbf{0.034--0.144} & \textbf{0.032--0.198} & \textbf{0.21--0.47} & \textbf{0.074--0.175} & $-$0.123--0.360 & \textbf{0.679--1.297} \\
			\midrule
			Copy-T & \textbf{0.008--0.203} & 0.026--0.724 & 0.39--0.87 & 0.010--0.048 & \textbf{$-$0.039--0.052} & \textbf{0.177--1.253} \\
			Copy-H & \textbf{0.075--0.140} & \textbf{0.156--0.265} & \textbf{0.28--0.36} & \textbf{0.032--0.046} & \textbf{$-$0.005--0.046} & \textbf{0.234--0.680} \\
			Copy-L & \textbf{0.027--0.197} & \textbf{0.061--0.165} & \textbf{0.57--0.71} & \textbf{0.024--0.053} & \textbf{$-$0.022--0.014} & \textbf{0.280--2.335} \\
			Grille & \textbf{0.490--0.596} & \textbf{0.173--0.255} & \textbf{0.59--0.67} & \textbf{0.032--0.045} & \textbf{0.026--0.093} & $-$0.010--0.017 \\
			HABIT & 0.009--2.405 & \textbf{0.093--0.658} & 0.16--0.74 & \textbf{0.026--0.096} & \textbf{$-$0.041--0.055} & \textbf{0.007--50.202} \\
			HABIT, no table & 0.029--0.885 & \textbf{0.070--0.985} & 0.22--0.87 & 0.013--0.076 & \textbf{$-$0.087--0.065} & \textbf{0.001--1.257} \\
						HABIT, all statistics & 0.009--0.805 & \textbf{0.062--0.485} & \textbf{0.16--0.69} & \textbf{0.028--0.080} & \textbf{$-$0.022--0.065} & $-$0.046--62.078 \\
			SPACE & 0.103--0.633 & \textbf{0.060--0.743} & 0.15--0.78 & 0.001--0.041 & \textbf{$-$0.361--0.133} & \textbf{0.022--0.679} \\
			SPACE, no table & 0.035--1.084 & \textbf{0.042--0.566} & \textbf{0.00--0.42} & 0.000--0.042 & 0.023--0.224 & \textbf{0.010--0.836} \\
			STRING & 0.077--0.677 & 0.010--0.320 & \textbf{0.23--0.60} & 0.003--0.031 & \textbf{$-$0.122--0.063} & $-$0.012--0.347 \\
			STRING, no crossing & 0.065--0.696 & 0.011--0.331 & \textbf{0.23--0.62} & 0.003--0.032 & \textbf{$-$0.115--0.065} & $-$0.015--0.331 \\
			STRING, no glyph rules & 0.056--1.053 & 0.010--0.374 & \textbf{0.00--0.25} & 0.004--0.070 & \textbf{$-$0.061--0.168} & $-$0.013--0.383 \\
			\bottomrule
		\end{tabular}

	\end{table*}
	
	All three programs reach the link, but none explains it. With the link held out, the texts of HABIT range from 0.009 to 2.405 bits, so they contain the 0.204 bits of the manuscript without predicting them. HABIT reaches the link also with the manuscript's tables for the two glyphs around a space switched off ($p = 0.24$ and 0.55), but its range stays wide (0.029 to 0.885 bits). SPACE predicts 0.103 to 0.633 bits, but only through the manuscript's odds of a space. With these odds switched off, its texts range from 0.035 to 1.084 bits, and the six measures of coupling separate them from the manuscript in both searches ($p = 0.01$ and 0.04). Fitted without the six measures of coupling and without any table of glyph pairs across spaces, STRING predicts a link of 0.077 to 0.677 bits, and these six measures do not separate its texts from the manuscript ($p = 0.79$ and 0.11). The same holds when its copied pieces never cross a space ($p = 0.81$ and 0.12) and when its chance of a space ignores the glyph classes ($p = 0.63$ and 0.09). No program passes the check on the fitted or on the 15 held-out statistics in any of the twelve fits ($p \le 0.05$). All three fail on the spaces in all twelve fits ($p = 0.01$), so the rules reject all three. In STRING the two glyphs around a space predict it far less well than in the manuscript (0.23 to 0.60 against 0.73). The manuscript writes the same string of glyphs as one word in one place and as two words in another, for 0.186 of its tokens beyond chance in ZL and 0.101 in IT (Table~\ref{tab:link}). Four of the 29 plaintexts that we enciphered in Section~\ref{sec:language}, in Latin, Italian and Middle French, reach 0.21 to 0.36. The central 95 percent of the texts of Copy-T, Copy-H, Copy-L, HABIT and STRING stay below 0.07, those of Grille below 0.10, and the texts of SPACE fall mostly below chance (median $-0.09$). When a word of ZL also appears split into two words, the part that splits off is most often one of s, ol and or at its start, or aiin at its end, and each of these parts attaches to many different words. On three measures that we fixed in advance, these parts fall inside the range of the plaintexts and also inside that of the copying texts, so the measures cannot tell whether they are function words. Every program also ties the endings of neighbouring words together more strongly than the manuscript does (0.027 bits, a value below all 29 plaintexts). Their medians lie between 0.19 and 0.31 bits, except those of STRING, which come closest at 0.073 to 0.080 bits, of Copy-L at 0.088 bits and of HABIT fitted to all statistics at 0.160 bits.
	
	The strength of the link comes from the glyphs. When the spaces of ZL are put back by word length alone, so that they fall without regard to the glyphs, the two glyphs at a space share 1.39 bits, twice as much as in any of the 29 plaintexts spaced the same way (0.22 to 0.69 bits). The glyphs of the manuscript are thus bound to their neighbours far more tightly than letters, as the low entropy of its glyph sequences also shows \citep{lindemann2020entropy,bowern2021linguistics}. Its real spaces keep 15 percent of this bond, about the average of the plaintexts (6 to 55 percent). The glyph pairs that the manuscript favours inside words are also those it favours across spaces. For the glyph pairs seen at least five times both inside words and across spaces, the pointwise mutual information inside words and that across spaces correlate at 0.49 in ZL, against $-0.09$ to 0.33 in the plaintexts. The central 95 percent of the texts of Copy-T, Copy-H and the nomenclators stay below 0.40, while the ranges of HABIT and SPACE, fitted without the measures of coupling, include the manuscript. The five scribes of \citet{davis2020scribes} favour different glyph pairs across spaces. These pairs agree more between two halves of one scribe's pages than between the pages of two scribes ($p = 0.005$, and also among the four scribes of Currier B alone). A space of the manuscript is also far easier to predict from the two glyphs around it than in any plaintext, and it says less about the glyphs beyond them. The uncertain spaces of ZL sit where the glyphs leave the decision open. The four glyphs around them give a probability of a space of 0.48, against 0.87 for certain spaces and 0.03 for gaps without one.
	
	We also measured the gaps on the scans. On the 171 scans that show a single page, we matched each line of ZL to its written line on the scan, with its certain spaces as anchors, and kept the 2,875 of 3,595 lines whose match passed our checks. Measured without the transliteration, the 82,672 gaps of at least three pixels in these lines come in two sizes, about 0.25 and 1.47 times the x-height of their line (two sizes fit better than one by 13,986 in the Bayesian information criterion). \citet{rozanova2026glyph} found the uncertain spaces narrower than the certain ones in word boxes drawn over the scans, and less clearly on six pages measured directly. Of the 1,635 uncertain spaces of the running text that we could measure, only one in four is as wide as a word space (95\% interval 22 to 28 percent). Inside words, glyph pairs that also often stand on either side of a space are written 1.8 times as far apart as pairs that almost never do. So on the page, too, the gap depends on the two glyphs around it. With its uncertain spaces removed (ZLj), ZL still links neighbouring words with 0.170 bits, more than any of the 29 plaintexts.
	
	Two neighbouring words stand at neighbouring places, so glyphs that prefer places in the line or in the paragraph \citep{feaster2022rightward} produce some link by themselves. To measure how much, we moved every word to a random slot of the same kind. A word kept the Currier dialect and section of its page, its place among the first or last three words of its line, the fifth of the line it stood in and the place of its line in the paragraph. Only 1.5 percent of neighbouring words stayed neighbours. The link of these texts is 0.014 bits in ZL, against 0.013 bits when words only stay within their dialect and section, so position gives about one percent of the link. What position leaves is 0.190 bits in ZL, 0.155 in ZLj and 0.174 in IT, more than in any of the 29 plaintexts treated the same way (0.025 to 0.138 bits). This holds under all 15 ways of grouping the slots that we tried, coarser or finer and within pages, sections or the whole text, under which position gives at most five percent of the link. It holds in the running text without the text written along circles and radii as well. The link is also stronger in Currier B. There it exceeds every plaintext in ZL, ZLj and IT (0.21 to 0.25 against at most 0.16 bits on the same pages), while in Currier A it does so only in ZL (0.136 against 0.129 bits) and stays within the range of the plaintexts in ZLj and IT.
	
	The link ends at the line break. The last glyph of a line and the first glyph of the next line of the same paragraph share no more than position gives ($-$0.001 bits, $p = 0.56$, and as little in ZLj and IT). In every plaintext the link runs on into the next line, at 0.68 to 1.30 times its strength inside lines ($p < 0.01$ in each). Even the lowest of these ratios would leave 0.13 bits at the line breaks of ZL, 36 times the standard deviation of the shuffled texts. The dependence between neighbouring words thus belongs to the line, which \citet{currier1976findings} already saw as a unit of the text. Of the programs in Table~\ref{tab:link}, only STRING and HABIT fitted to all statistics reach the link inside lines and end it at the line break as the manuscript does. STRING starts every line with a new string of glyphs, so no copied string runs on into the next line. A line break in its texts keeps $-$0.01 to 0.35 of the link inside lines, against 0.18 to 1.25 in Copy-T, 0.23 to 0.68 in Copy-H, 0.28 to 2.34 in Copy-L and 0.02 to 0.68 in SPACE. HABIT does both only in its fits to all statistics, whose texts keep anything from nothing to 62 times the link at a line break. Grille ends the link at the line break too, but links neighbouring words in a line 2.5 to 3 times as strongly as the manuscript. In Copy-W, Copy-N and every cipher the link beyond position stays below 0.07 bits in the central 95 percent of the texts, against 0.19 in ZL.
	
	\section{Traces of Language That Need No Decipherment}
	\label{sec:language}
	
	Word statistics alone leave both accounts standing. We therefore look for a property that real language gives a text whatever the script, and that a word code keeps.
	
	\subsection{Word pairs in a line}
	
	In a text with meaning, words that go with each other, such as a noun and its verb, or the parts of a stock phrase, share a line more often than chance would place them there, and they do so again in other parts of the text. Collocation measures capture the first part of this pattern \citep{church1990association,dunning1993accurate}, and our measure adds the second. A word code keeps this pattern, since every plaintext word keeps its codeword or one of its homophones. A copying text has local pairs too, but a pair that arises from copying on one page has no reason to recur on pages written elsewhere.
	
	We measure the pattern as follows. For the 300 most frequent word types we count, for every pair of types, the lines that contain both. We compare each count with its mean over 20 versions of the text with words shuffled inside each page, keeping the first and last word of every line in place, and divide the excess by $\sqrt{\text{expected count}+1/2}$. Pairs of equal types and pairs that become equal when one glyph is removed from each are left out, so that copying and near-copying cannot produce the signal. We then split the text into two halves of alternating blocks of ten pages, compute the excess of every pair in each half, remove the overall tendency of each word to share lines, and take the Spearman correlation of the two halves over all pairs expected at least once in each half. We remove the tendency of each word by fitting the excess of every pair as the sum of one value for each of its two words, by least squares, and subtracting this fit. We call the correlation the \textit{word-pair replication}. A pattern that recurs across the book gives a high value, and pairs that arise locally give a value near zero. The mean square of the excess, which we call \textit{overdispersion}, shows how strongly pairs deviate from chance at all, and the page replication and overdispersion, measured on pages instead of lines, show how pairs cluster. Variants with pairs at least two words apart and with the 150 or 600 top types give the same picture (Appendix~\ref{app:pairs}). A joint check compares a group of texts with the manuscript on seven of these statistics, with the Mahalanobis distance of Section~\ref{sec:fitting}. They are the replication in lines, for pairs at least two words apart and on pages, the overdispersion in lines and on pages, the drop in replication when the line breaks are moved by half a line, and the excess of pairs two glyphs apart over other pairs.
	
	\begin{figure}[t]
		\centering
		\includegraphics[width=0.9\columnwidth]{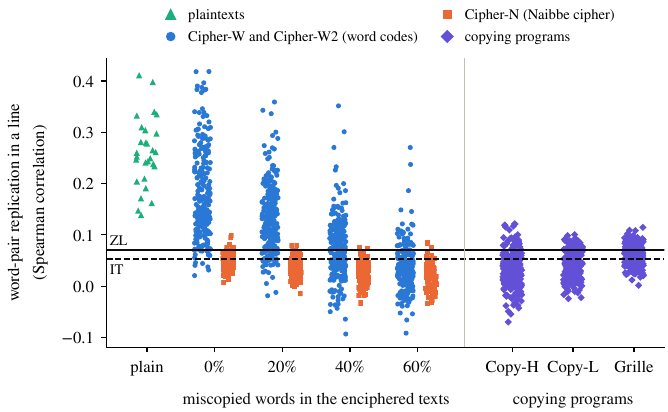}
		\caption{Word-pair replication of 28 natural texts in 16 languages, and of word codes (Cipher-W and Cipher-W2) and Cipher-N of held-out texts in the same languages at four rates of miscopying, and of the three copying programs of Section~\ref{sec:morecopying}, against ZL (solid line) and IT (dashed line).}
		\label{fig:pairs}
	\end{figure}
	
	\paragraph{Natural texts}
	We cut 28 stretches of 37,554 words from held-out documents in 16 languages and laid each into the pages and lines of ZL. Their word-pair replication lies between 0.139 and 0.411. The manuscript reaches 0.071 in ZL and 0.053 in IT (Figure~\ref{fig:pairs}). It lies below all 28 texts, and so do its replication for pairs at least two words apart and its overdispersion. Hebrew, Arabic and Persian, written as consonants, spell many words alike yet reach 0.147 to 0.411.
	
	\paragraph{Ciphers of held-out texts}
	We enciphered 29 plaintexts, 27 stretches drawn again from the held-out texts of the 16 languages and two of abbreviated Latin, with the best fits of every cipher program, at the fitted rate of miscopying and at 0, 20, 40 and 60 percent (Table~\ref{tab:pairs}). The word and syllable codes keep most of the pattern of their plaintext. Averaged over the languages, each program and fit keeps 52 to 84 percent at the fitted rate of miscopying and 11 to 23 percent at 60 percent miscopying. At the fitted rate only 5 to 9 of about a hundred texts of each word or syllable code lie at or below ZL. These come from plaintexts that show the pattern weakly to begin with, above all abbreviated Latin, which reaches only 0.102 before enciphering, and the troubadour poems of Old Occitan. The joint check cannot separate ZL from the word and syllable codes at the fitted rate ($p = 0.32$ to 0.61), while it separates every group at 60 percent miscopying ($p = 0.02$).
	
	Two programs behave differently. Cipher-N removes most of the pattern even without miscopying, because it splits words into letters and letter pairs whose homophones the scribe picks from what he sees. Its texts lie at or below ZL in 102 of 108 cases. They differ from the manuscript on pages. Their word pairs recur on pages more strongly, with a page replication of 0.050 and 0.058 against 0.024 in ZL. None of the 108 texts comes down to ZL on this measure, and the joint check separates every setting ($p = 0.02$). The gap stays when the plaintext is written page by page, each page a stretch of another document as in a herbal, and all 64 Cipher-N texts of such plaintexts lie above ZL. Cipher-C shows almost no recurring pairs with the two classes of its first fit and matches the line statistics of ZL with the 747 classes of its second. The clustering of its pairs on pages misses that of the manuscript in both fits, far too strong with two classes and too weak with 747, and the joint check separates both from ZL ($p = 0.02$).
	
	The nomenclators lose more of the pattern than word codes. Four of their six fits miscopy 37 to 56 percent of words, and every rare word is spelled out in several written words. At the fitted rate of miscopying, 74 of 116 texts of Cipher-M, 96 of 116 of Cipher-M2 and 103 of 108 of Cipher-MN lie at or below ZL. The joint check separates Cipher-M, Cipher-MN and the first fit of Cipher-M2 from ZL (each $p = 0.02$). That fit spells nearly every word, and like Cipher-N its pairs recur on pages too strongly, with a page replication of 0.106 against 0.024. The second fit of Cipher-M2 cannot be separated from ZL ($p = 0.85$) or from IT ($p = 0.17$). Without miscopying, the same code keeps the pattern far above the manuscript, with 9 of 58 texts at or below ZL (joint $p = 0.02$). A nomenclator could leave one more trace. It spells a rare word with one written word per letter, so when the rare word comes back many pages later, a whole sequence of written words comes back with it. Copying repeats what is near. By the rule we fixed, sequences of three words that come back ten or more pages later, with one word allowed to differ, do not separate the accounts. ZL has more of them than the central 95 percent of the texts of Copy-T, Copy-H and SPACE, but it lies inside the range of all three nomenclators, of the plaintexts and of HABIT (Table~\ref{tab:returns}).
	
	The three copying programs of Section~\ref{sec:morecopying} predict word-pair replications between 0.022 and 0.065 on average, close to ZL (Figure~\ref{fig:pairs}), but none of them matches the overdispersion of pairs in lines or on pages, and the joint check separates every fit ($p = 0.01$).
	
	\begin{table}[t]
		\centering
		\small
		\setlength{\tabcolsep}{4pt}
				\caption{Enciphered texts whose word-pair replication lies at or below that of ZL, for the plaintexts of 16 languages and abbreviated Latin, both fits and two replicates. The fitted rate of miscopying, taken from the best parameter set of each fit, lies between 0.4 and 15 percent for the word and syllable codes, at 5 and 9 percent for Cipher-N, at 25 and 54 percent for Cipher-C, at 50 and 49 percent for Cipher-M, at 37 and 56 percent for Cipher-M2 and at 3 and 0.1 percent for Cipher-MN. Cipher-N, Cipher-S and Cipher-MN take plaintexts in plain letters only and leave out abbreviated Latin.}
		\label{tab:pairs}
		\begin{tabular}{@{}lrrrrrr@{}}
			\toprule
			& & \multicolumn{5}{c}{Miscopied words} \\
			\cmidrule(l){3-7}
			Program & Texts & fitted & 0\% & 20\% & 40\% & 60\% \\
			\midrule
			Cipher-W & 116 & 8 & 9 & 23 & 58 & 102 \\
			Cipher-W2 & 116 & 5 & 6 & 20 & 53 & 89 \\
			Cipher-S & 108 & 9 & 6 & 22 & 65 & 101 \\
			Cipher-C & 116 & 88 & 59 & 68 & 76 & 91 \\
			Cipher-N & 108 & 102 & 102 & 106 & 106 & 106 \\
			Cipher-M & 116 & 74 & 23 & 38 & 58 & 93 \\
			Cipher-M2 & 116 & 96 & 64 & 74 & 85 & 96 \\
			Cipher-MN & 108 & 103 & 97 & 107 & 107 & 105 \\
			\bottomrule
		\end{tabular}

	\end{table}
	
	\paragraph{Where the manuscript stands}
	A plaintext that shows the pattern at the strength of an ordinary text loses it under a word code only if a large share of its words is miscopied. Such a word code no longer looks like the manuscript in other ways. The enciphered controls of Section~\ref{sec:wordcodes} with 60 percent miscopying all lie at a distance of at least 0.73 from the manuscript on the 22 fitted statistics, against 0.48 and 0.50 for the closest simulations of Copy-T and Cipher-W (Table~\ref{tab:results}). A word code matches the word pairs of the manuscript only if its plaintext is a text like abbreviated Latin, whose word pairs recur weakly. Most of the best simulations of Cipher-W and Cipher-W2 in Section~\ref{sec:results} chose exactly this plaintext. Section~\ref{sec:wordcodes} tests these codes directly. The second fit of Cipher-M2 is the exception. With ordinary plaintexts it matches the word pairs of the manuscript and passes the held-out check, although its scribe miscopies more than half of its words. Its predictive texts lie at a median distance of 0.74 from the manuscript, about as far as the controls with 60 percent miscopying, but the closest lies at 0.58. Section~\ref{sec:nomen} tests it directly.
	
	\paragraph{Which pages carry the pattern}
	The manuscript changes its vocabulary between Currier A and B, so a weak replication could come from comparing halves that hold different shares of the two dialects. Alternating the blocks within each dialect, using only A pages or only B pages, or using A and B as the two halves gives the same picture. Every plaintext except abbreviated Latin lies above ZL in every view, apart from the Church Slavonic texts on the A pages alone. Removing the near copies, which might be filler words hiding a real text, does not change the result either (Section~\ref{sec:other}).
	
	\subsection{Natural languages in invented scripts}
	
	The manuscript could be a natural language written in a newly invented script, without any cipher. We wrote held-out stretches of all 16 languages into the layout of ZL in four kinds of script, an alphabet with one sign per letter, a positional script with separate signs for a letter at the start, middle and end of a word, a syllabary, and a consonantal script that writes vowels only at the start of a word. These 64 texts lie at distances between 1.23 and 154 from the manuscript on the 22 fitted statistics, against 0.25 for IT and 0.48 for the closest simulation of Copy-T. The closest is Italian in a consonantal script. Leaving out the four layout statistics, we compared 55 statistics, the 22 fitted and 15 held-out ones, the six measures of coupling and the six shares of copying choices of Section~\ref{sec:morecopying}, and six measures from the solvers of Section~\ref{sec:decipher}. The manuscript lies outside the range of all 64 texts on 11 of them. Among them, it repeats the previous word more often than any of them (a log ratio to the page-shuffled text of 0.185 against at most $-0.427$), its adjacent words share less information (0.116 against at least 0.127 bits), its words differ more between Currier A and B (0.467 against at most 0.434 bits), and the frequency of its words correlates more strongly with their number of one-glyph neighbours (0.587 against at most 0.533). No script we tried turns a natural text into something close to the manuscript.
	
	\subsection{A gibberish classifier, calibrated}
	
	\citet{gaskell2022gibberish} trained a classifier on gibberish written by volunteers and on meaningful texts and found the manuscript closer to gibberish. We trained the same kind of random forest on their 38 samples of gibberish and on 102 prose texts in five languages. It separates the two with an out-of-bag AUC of 0.985 and calls the manuscript gibberish with probability 0.71. It also calls gibberish 53 of 54 meaningful texts enciphered with Cipher-W and Cipher-W2, with mean probabilities between 0.55 and 0.62, and all five enciphered with Cipher-N. A text that looks like gibberish to this classifier can still be a cipher, so the classifier cannot tell the two accounts apart.
	
	\subsection{Published measures of language}
	\label{sec:published}
	
		\begin{table}[t]
		\centering
		\small
		\setlength{\tabcolsep}{3.5pt}
				\caption{Published measures of language on the manuscript (ZL) and on texts whose answer we know. Table~\ref{tab:published_entropy} gives the conditional entropy. The classifier is that of \citet{lee2010pictish}, with its two ratios $C_r$ and $U_r$ (Appendix~\ref{app:stats}) and the entropy that $U_r$ divides. Repetition is the share of repeated units that repeat the unit just before them, within a word, line or paragraph \citep{sproat2014statistical}, and a text above 0.10 counts as nonlinguistic. Plaintexts gives the range of the 29 plaintexts, whose glyphs are letters, or how many of them lie in a class. Ciphers counts the 16 cipher groups, in parentheses those without miscopying, and copying the 13 copying groups, whose central 95 percent contains the value of ZL or at least half of whose texts share its class. $^{\dagger}$Copy-T is not among them. Overlap counts the copying groups whose central 95 percent overlaps the range of the plaintexts, in parentheses without the two of abbreviated Latin where that differs. Bold marks plaintexts that do not reach ZL. ZLj and IT are reached on every value of both tables as well, and the classifier calls the words of 97 percent of all texts words.}
		\label{tab:published}
		\begin{tabular}{@{}llrrrrr@{}}
			\toprule
			& & & & \multicolumn{2}{c}{Groups reaching ZL} & \\
			\cmidrule(l){5-6}
			& Computed on & ZL & Plaintexts & Ciphers & Copying & Overlap \\
			\midrule
			\multicolumn{7}{@{}l}{\textit{Classifier}} \\
			class & glyphs & letters & all letters & 16 (8) & 13 &  \\
			& words & words & all words & 16 (8) & 13 &  \\
			$C_r$ & glyphs & 16.910 & 15.459--29.179 & 9 (6) & 10$^{\dagger}$ & 13 (12) \\
			& words & 8.852 & \textbf{6.551--8.819} & 8 (3) & 7 & 12 \\
			$U_r$ & glyphs & 0.625 & \textbf{0.791--0.868} & 5 (2) & 9 & 6 \\
			& words & 2.332 & 2.035--3.041 & 10 (3) & 10 & 12 \\
			entropy (bits) & glyphs & 2.550 & \textbf{3.345--3.991} & 3 (3) & 7 & 6 \\
			& words & 4.791 & 3.558--4.892 & 13 (5) & 7 & 13 \\
			\addlinespace
			\multicolumn{7}{@{}l}{\textit{Repetition}} \\
			share & glyphs in words & 0.591 & \textbf{0.053--0.442} & 12 (6) & 9 & 9 \\
			& glyphs in lines & 0.095 & 0.016--0.095 & 9 (5) & 11 & 11 \\
			& words in lines & 0.201 & \textbf{0.001--0.163} & 10 (4) & 11$^{\dagger}$ & 8 (0) \\
			& words in paragraphs & 0.050 & 0.000--0.050 & 10 (4) & 11$^{\dagger}$ & 11 (0) \\
			side of 0.10 & glyphs in words & above & 24 of 29 above & 16 (8) & 13 &  \\
			& glyphs in lines & below & none above & 1 (0) & 4 &  \\
			& words in lines & above & \textbf{2 of 29 above} & 11 (3) & 13 &  \\
			& words in paragraphs & below & none above & 14 (7) & 11$^{\dagger}$ &  \\
			\addlinespace
			\multicolumn{7}{@{}l}{\textit{Yule's K}} \\
			& glyphs & 812 & 601--826 & 15 (7) & 6$^{\dagger}$ & 10 \\
			& words & 33.2 & 16.2--165.5 & 14 (7) & 11 & 13 \\
			\bottomrule
		\end{tabular}

	\end{table}
	
	\begin{table}[p]
		\centering
		\small
		\setlength{\tabcolsep}{3.5pt}
				\caption{Conditional entropy of a unit given the unit before it \citep{rao2009entropic} on the manuscript (ZL) and on texts whose answer we know, for all glyphs or the 20 most frequent and for the 20 to 400 most frequent words. Columns as in Table~\ref{tab:published}.}
		\label{tab:published_entropy}
		\begin{tabular}{@{}lrrrrr@{}}
			\toprule
			& & & \multicolumn{2}{c}{Groups reaching ZL} & \\
			\cmidrule(l){4-5}
			Computed on & ZL & Plaintexts & Ciphers & Copying & Overlap \\
			\midrule
			\multicolumn{6}{@{}l}{\textit{Bits}} \\
			all glyphs in words & 2.233 & \textbf{2.764--3.721} & 7 (5) & 8 & 10 \\
			top 20 glyphs in words & 2.196 & \textbf{2.736--3.674} & 9 (5) & 8 & 9 \\
			all glyphs in lines & 2.255 & \textbf{3.012--3.654} & 6 (5) & 7 & 7 \\
			top 20 glyphs in lines & 2.206 & \textbf{2.971--3.575} & 8 (5) & 7 & 7 \\
			top 20 words in lines & 3.641 & \textbf{1.687--3.463} & 11 (6) & 11 & 11 \\
			top 40 words in lines & 4.663 & \textbf{2.519--4.275} & 9 (5) & 8 & 11 \\
			top 60 words in lines & 5.238 & \textbf{3.040--4.772} & 10 (6) & 7 & 11 \\
			top 80 words in lines & 5.598 & \textbf{3.336--5.205} & 12 (6) & 6 & 11 \\
			top 100 words in lines & 5.869 & \textbf{3.597--5.472} & 13 (7) & 6 & 11 \\
			top 120 words in lines & 6.071 & \textbf{3.791--5.687} & 14 (8) & 6 & 11 \\
			top 140 words in lines & 6.250 & \textbf{3.990--5.863} & 15 (8) & 6 & 11 \\
			top 160 words in lines & 6.398 & \textbf{4.109--5.959} & 15 (8) & 6 & 11 \\
			top 180 words in lines & 6.522 & \textbf{4.222--6.096} & 15 (8) & 7 & 11 \\
			top 200 words in lines & 6.651 & \textbf{4.342--6.209} & 15 (8) & 7 & 11 \\
			top 220 words in lines & 6.752 & \textbf{4.462--6.307} & 15 (8) & 7 & 11 \\
			top 240 words in lines & 6.845 & \textbf{4.571--6.422} & 15 (8) & 7 & 11 \\
			top 260 words in lines & 6.921 & \textbf{4.639--6.506} & 14 (7) & 7 & 11 \\
			top 280 words in lines & 6.994 & \textbf{4.705--6.581} & 14 (7) & 7 & 11 \\
			top 300 words in lines & 7.059 & \textbf{4.778--6.648} & 13 (7) & 8 & 11 \\
			top 320 words in lines & 7.125 & \textbf{4.844--6.713} & 13 (7) & 8 & 11 \\
			top 340 words in lines & 7.186 & \textbf{4.915--6.779} & 13 (7) & 9 & 11 \\
			top 360 words in lines & 7.240 & \textbf{4.965--6.835} & 13 (7) & 9 & 11 \\
			top 380 words in lines & 7.292 & \textbf{5.011--6.895} & 13 (7) & 9 & 11 \\
			top 400 words in lines & 7.338 & \textbf{5.043--6.933} & 13 (7) & 9 & 11 \\
			\addlinespace
			\multicolumn{6}{@{}l}{\textit{Divided by the logarithm of the number of units kept}} \\
			all glyphs in words & 0.451 & \textbf{0.611--0.834} & 7 (5) & 7 & 7 \\
			top 20 glyphs in words & 0.508 & \textbf{0.633--0.850} & 9 (5) & 8 & 9 \\
			all glyphs in lines & 0.451 & \textbf{0.657--0.808} & 6 (5) & 6 & 6 \\
			top 20 glyphs in lines & 0.510 & \textbf{0.687--0.827} & 8 (5) & 7 & 7 \\
			top 20 words in lines & 0.842 & \textbf{0.390--0.801} & 11 (6) & 11 & 11 \\
			top 40 words in lines & 0.876 & \textbf{0.473--0.803} & 9 (5) & 8 & 11 \\
			top 60 words in lines & 0.887 & \textbf{0.515--0.808} & 10 (6) & 7 & 11 \\
			top 80 words in lines & 0.886 & \textbf{0.528--0.823} & 12 (6) & 6 & 11 \\
			top 100 words in lines & 0.883 & \textbf{0.541--0.824} & 13 (7) & 6 & 11 \\
			top 120 words in lines & 0.879 & \textbf{0.549--0.823} & 14 (8) & 6 & 11 \\
			top 140 words in lines & 0.877 & \textbf{0.560--0.822} & 15 (8) & 6 & 11 \\
			top 160 words in lines & 0.874 & \textbf{0.561--0.814} & 15 (8) & 6 & 11 \\
			top 180 words in lines & 0.871 & \textbf{0.564--0.814} & 15 (8) & 7 & 11 \\
			top 200 words in lines & 0.870 & \textbf{0.568--0.812} & 15 (8) & 7 & 11 \\
			top 220 words in lines & 0.868 & \textbf{0.573--0.810} & 15 (8) & 7 & 11 \\
			top 240 words in lines & 0.866 & \textbf{0.578--0.812} & 15 (8) & 7 & 11 \\
			top 260 words in lines & 0.863 & \textbf{0.578--0.811} & 14 (7) & 7 & 11 \\
			top 280 words in lines & 0.860 & \textbf{0.579--0.810} & 14 (7) & 7 & 11 \\
			top 300 words in lines & 0.858 & \textbf{0.581--0.808} & 13 (7) & 8 & 11 \\
			top 320 words in lines & 0.856 & \textbf{0.582--0.807} & 13 (7) & 8 & 11 \\
			top 340 words in lines & 0.855 & \textbf{0.584--0.806} & 13 (7) & 9 & 11 \\
			top 360 words in lines & 0.853 & \textbf{0.585--0.805} & 13 (7) & 9 & 11 \\
			top 380 words in lines & 0.851 & \textbf{0.585--0.805} & 13 (7) & 9 & 11 \\
			top 400 words in lines & 0.849 & \textbf{0.583--0.802} & 13 (7) & 9 & 11 \\
			\bottomrule
		\end{tabular}

	\end{table}
	
	Four of the measures used to decide whether a script writes a language (Section~\ref{sec:related}) can be computed on any text. They are the conditional entropy of a sign given the sign before it \citep{rao2009entropic}, the entropy classifier of \citet{lee2010pictish}, the share of repeated signs that repeat the sign just before them \citep{sproat2014statistical} and Yule's K, which grows as a text repeats its units more \citep{tanakaishii2015constancy}. We computed them on the glyphs and on the words of the manuscript and of 5,497 texts whose answer we know. These are the 29 plaintexts, 1,468 ciphers of the held-out texts of Section~\ref{sec:language} and 4,000 predictive texts of the copying programs. The ciphers come from eight programs, each at its fitted rate of miscopying and without miscopying, which gives 16 groups. The copying texts come from nine programs and from four variants of Section~\ref{sec:link} that switch off one of their parts, which gives 13 groups. The glyphs of a plaintext are its letters. For the conditional entropy of words we keep the 20 to 400 most frequent words, the range that \citet{rao2009entropic} plot. With all settings the manuscript has 66 values (Appendix~\ref{app:stats}). We fixed the rule for a verdict before the run. A group reaches a value of the manuscript if the value lies between the smallest and the largest of the plaintexts, or inside the central 95 percent of the texts of a program.
	
	No measure decides (Tables~\ref{tab:published} and~\ref{tab:published_entropy}). Every one of the 66 values is reached by at least one copying group and by texts with language, and the same holds for ZLj and IT. Texts with language that were never miscopied reach every value as well, either the plaintexts or a cipher without miscopying. Copy-T alone reaches 61 of the 66. The classifier calls the glyphs of all 5,497 texts letters and the words of 97 percent of them words, whether they have a language or not. Within lines the repetition share of the manuscript is 0.20, above the threshold of 0.10, so the measure calls it nonlinguistic. Most texts in all 13 copying groups and in 11 of the 16 cipher groups also lie above it, three of these groups without miscopying, while among the plaintexts only the two of abbreviated Latin do. Within paragraphs the manuscript stays below the threshold, and so do most texts in 11 copying groups. For glyphs in words the threshold calls 24 of the 29 plaintexts nonlinguistic.
	
	On every numerical value the range of the plaintexts overlaps that of at least six copying groups. Only the repetition of words separates them once abbreviated Latin is set aside. Within lines and within paragraphs, all other plaintexts lie below the central 95 percent of every copying group. The manuscript lies among the copying texts on both values, but also among the ciphers of 10 of the 16 groups, four of them without miscopying.
	
	The manuscript lies outside the range of the plaintexts on 53 of the 60 numerical values. Its glyphs, for instance, are easier to predict from the glyph before them than the letters of any plaintext, as earlier studies found \citep{lindemann2020entropy,bowern2021linguistics}. Ciphers and copying programs both reach these values. The measures respond to how regular a text is, and both accounts make a text as regular as the manuscript, so the measures cannot tell the two apart.
	
	\section{Decipherment With Calibrated Controls}
	\label{sec:decipher}
	
	A cipher that fits the manuscript makes a claim that can be tested directly. If the manuscript is such a cipher, a solver that reads texts made by the cipher should also read the manuscript. We test four families in turn, each checked on controls whose plaintext we know.
	
	\subsection{Letter ciphers}
	\label{sec:reading}
	
	The best fit of Cipher-N claims that every frequent word type stands for one plaintext letter or letter pair, and that several types stand for the same unit.
	
	\paragraph{Solver}
	We read the 800 most frequent word types of a text, which cover 74\% of the tokens of ZL. Each type $c$ receives a reading $m(c)$ among the 383 to 543 Naibbe units of a language. Rarer types and page boundaries break the text into runs. Under a bigram model $P$ of the language the log-likelihood of a reading is
	\begin{equation}
		\begin{split}
			J(m)={}&\sum_{\text{runs}}\Bigl(\log P\bigl(m(c_1)\bigr)\\
			&\qquad+\sum_{t\ge 2}\log P\bigl(m(c_t)\mid m(c_{t-1})\bigr)\Bigr)\\
			&+\sum_{c} n_c \log\frac{n_c}{Q\bigl(m(c)\bigr)},
		\end{split}
	\end{equation}
	where $c_1, c_2, \ldots$ are the types of a run, $n_c$ is the count of type $c$ and $Q(u)$ is the number of tokens read as unit $u$. The last sum is the log-probability of the homophone choices. It is zero when a type has a reading of its own and negative when types share a reading, so the solver joins types only when the context supports it. A type may only take a reading whose frequency in the language lies between a third and a hundred times its own. We maximise $J$ by simulated annealing \citep{kirkpatrick1983annealing} from four starting points. The language models use absolute discounting \citep{ney1994structuring} and see none of the control texts.
	
	\paragraph{Controls}
	The positive controls are held-out texts of 40,000 words in each of the 16 languages, enciphered by the best Cipher-N of each search with its scribe, nulls and miscopies. We solve every text twice, as it is and with its words shuffled inside each page, and the manuscript also with its words shuffled only among words of the same line position, which records whether a word opens, ends or sits inside its line and whether its line opens a paragraph. The gap is the best $J$ per solved token of a text minus that of its shuffled version. We also solve 20 predictive texts of Cipher-N and of Copy-T, which gives each a predicted range of gaps.
	
		\begin{table*}[t]
		\centering
		\scriptsize
		\setlength{\tabcolsep}{2pt}
				\caption{Reading test for letter ciphers. Gaps are in bits per solved token against the page-shuffled version, and for the manuscript also against the version shuffled within line positions (second value). Accuracy is the share of correctly read tokens among the tokens of the 800 solved types. Accuracy and gap give the two controls of the model's language, other languages the mean gap of the 30 controls of the other languages, and under other models the mean gap of the two controls under the other 15 models. The program columns give the range of 20 predictive texts, and the last row averages each text over the 16 models. Leaving out Hebrew, every control has its largest scaled gap under the model of its own language except one of Old French, which peaks under the Middle French model. On the raw gaps one of Arabic also peaks elsewhere, under the Greek model. All 20 texts of Cipher-N peak under the model of their plaintext language.}
		\label{tab:reading}
		\begin{tabular}{@{}lcccccccc@{}}
			\toprule
			& \multicolumn{4}{c}{Controls} & \multicolumn{2}{c}{Program texts} & \multicolumn{2}{c}{Manuscript} \\
			\cmidrule(lr){2-5}\cmidrule(lr){6-7}\cmidrule(lr){8-9}
			& accuracy & & other & under other & & & & \\
			Language model & (\%) & gap & languages & models & Cipher-N & Copy-T & ZL & IT \\
			\midrule
						Arabic & 40 / 33 & 0.209 / 0.174 & 0.145 & 0.151 / 0.144 & 0.080--0.190 & $-$0.040--0.066 & 0.107 / 0.102 & 0.098 / 0.102 \\
			Church Slavonic & 66 / 62 & 0.543 / 0.512 & 0.267 & 0.272 / 0.259 & 0.192--0.365 & 0.008--0.088 & 0.147 / 0.148 & 0.134 / 0.149 \\
			German & 76 / 73 & 0.628 / 0.557 & 0.253 & 0.294 / 0.254 & 0.192--0.556 & 0.004--0.114 & 0.173 / 0.182 & 0.143 / 0.153 \\
			Middle English & 76 / 73 & 0.471 / 0.446 & 0.265 & 0.229 / 0.217 & 0.166--0.352 & 0.012--0.100 & 0.169 / 0.187 & 0.142 / 0.156 \\
			Persian & 42 / 41 & 0.238 / 0.219 & 0.141 & 0.158 / 0.153 & 0.081--0.200 & $-$0.040--0.064 & 0.082 / 0.082 & 0.056 / 0.055 \\
			Middle French & 85 / 84 & 0.642 / 0.645 & 0.285 & 0.292 / 0.296 & 0.195--0.418 & 0.015--0.110 & 0.156 / 0.164 & 0.133 / 0.143 \\
			Old French & 74 / 73 & 0.451 / 0.429 & 0.256 & 0.288 / 0.261 & 0.160--0.381 & 0.005--0.080 & 0.151 / 0.157 & 0.121 / 0.124 \\
			Greek & 74 / 73 & 0.676 / 0.648 & 0.339 & 0.301 / 0.284 & 0.249--0.516 & $-$0.047--0.109 & 0.236 / 0.241 & 0.210 / 0.211 \\
			Hebrew & 6 / 1 & 0.013 / 0.013 & 0.080 & 0.040 / 0.047 & 0.049--0.107 & $-$0.004--0.062 & 0.042 / 0.046 & 0.038 / 0.036 \\
			Italian & 79 / 76 & 0.611 / 0.578 & 0.295 & 0.302 / 0.295 & 0.221--0.726 & 0.012--0.108 & 0.174 / 0.186 & 0.157 / 0.144 \\
			Latin & 85 / 82 & 0.573 / 0.541 & 0.248 & 0.272 / 0.265 & 0.211--0.635 & $-$0.018--0.087 & 0.151 / 0.152 & 0.124 / 0.136 \\
			Old Norse & 73 / 69 & 0.431 / 0.398 & 0.228 & 0.230 / 0.219 & 0.148--0.353 & 0.006--0.104 & 0.131 / 0.138 & 0.136 / 0.135 \\
			Czech & 68 / 69 & 0.402 / 0.395 & 0.237 & 0.224 / 0.211 & 0.184--0.357 & 0.005--0.084 & 0.141 / 0.145 & 0.129 / 0.135 \\
			Old Spanish & 78 / 77 & 0.655 / 0.613 & 0.285 & 0.348 / 0.329 & 0.212--0.502 & $-$0.004--0.099 & 0.179 / 0.192 & 0.155 / 0.153 \\
			Portuguese & 85 / 83 & 0.673 / 0.614 & 0.275 & 0.316 / 0.288 & 0.194--0.477 & 0.005--0.109 & 0.145 / 0.149 & 0.129 / 0.128 \\
			Old Occitan & 75 / 73 & 0.494 / 0.484 & 0.279 & 0.262 / 0.252 & 0.198--0.424 & $-$0.003--0.073 & 0.180 / 0.183 & 0.150 / 0.151 \\
			\addlinespace
			Mean of 16 models &  &  &  &  & 0.175--0.383 & 0.008--0.082 & 0.148 / 0.153 & 0.128 / 0.132 \\
			\bottomrule
		\end{tabular}

	\end{table*}
	
\begin{figure*}[t]
		\centering
		\includegraphics[width=0.9\textwidth]{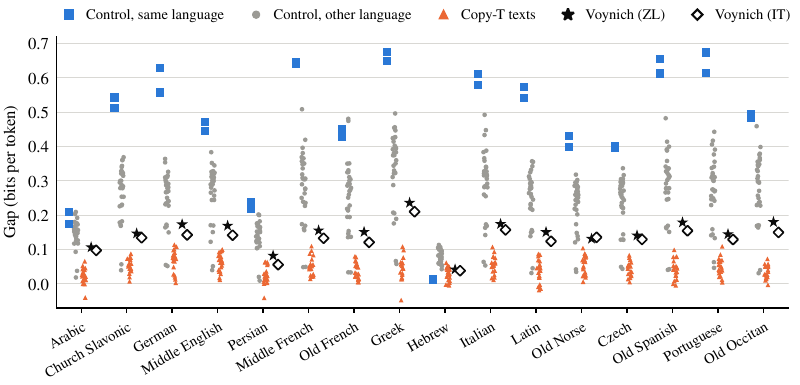}
		\caption{Gap between each text and its page-shuffled version under each of the 16 language models. Except under the Hebrew model, whose controls the solver cannot read, both transliterations fall below the two controls of the model's own language, and ZL stays above all texts of Copy-T.}
		\label{fig:reading}
	\end{figure*}

	\paragraph{Results}
	The solver reads the controls of 15 of the 16 languages (Table~\ref{tab:reading}, Figure~\ref{fig:reading}). It reads 33\% to 85\% of their solved tokens correctly. The controls of other languages reach larger gaps under some models than under others, most of all under the Greek model. We therefore also scale each gap by the mean and standard deviation of the gaps that the controls of the other languages reach under the same model. On this scale 29 of the 30 controls of these 15 languages and all 20 texts of Cipher-N peak under the language of their plaintext, against 28 of the 30 controls on the raw gaps. The one exception on this scale is an Old French control, which peaks under the Middle French model. The solver reads only 1\% and 6\% of the solved tokens of the two Hebrew controls correctly, so the test is undecided for Hebrew. A longer search does not change this. With four times as many restarts or up to 16 times longer cooling, the solver finds keys that score about as high as the key it reaches when started from the true key, yet it still reads at most 10\% of the tokens correctly. Under the Hebrew model, wrong keys fit these controls as well as the right one. Under each of the other 15 models, ZL and IT lie below both controls of the model's language, and ZL reaches at most 0.61 times the smaller of their gaps. ZL also lies above all 20 texts of Copy-T under each of these models. Its gap ranges from 0.082 to 0.236 under them and that of IT from 0.056 to 0.210. Read with the models of other languages, the controls of Arabic and Persian show about as much order as ZL, so a letter cipher of a language we did not model could look like the manuscript. Averaged over all 16 models, ZL reaches 0.148. All 20 texts of Cipher-N lie above it, and all 20 texts of Copy-T lie below it. ZL lies below the mean of the controls of the other languages under each of the 16 models. The readings of ZL under the five models of the main corpus are strings of letters without recognisable words. A letter cipher of this kind does not fit the manuscript in any of the 15 languages whose controls the solver reads.
	
	\paragraph{Where the structure comes from}
	The manuscript lies between the copying texts and the letter cipher. In separate runs with only the five models of the main corpus, where the gap of ZL averages 0.14 to 0.15, Cipher-W, Cipher-W2, Cipher-S and Cipher-C reach it as well. To find out what the solver picks up, we shuffled the manuscript in these runs in ways that keep one kind of structure each. A shuffle that swaps only words with the same first and the same last glyph keeps the last glyph of every word next to the first glyph of the following word, and it keeps 78\% of the gap of ZL. Swaps inside random groups of the same sizes keep nothing. Shuffles that keep only the first or only the last glyph of every word, its length, its line or its line position keep almost none of it. In the texts of Cipher-N the first shuffle leaves 9\% of the gap and in those of Cipher-W and Cipher-W2 12\%. The link of the manuscript is 0.204 bits, against 0.078 in Copy-T and 0.014 in these ciphers. A version of Copy-T that we forced to match this link reaches about half of the gap of ZL (0.079 against 0.144). So most of the structure that the solver finds sits at the boundaries between words, the same coupling that \citet{schwartz2026voynich}, \citet{rozanova2026glyph} and \citet{parisel2026layered} report, and a copying scribe with a habit at word boundaries explains half of it.
	
	\subsection{Word codes and miscopying}
	\label{sec:wordcodes}
	
	The letter solver cannot test a word code, because a word code puts the grammar of its plaintext into the order of whole words. We built a solver for word codes that learns the inverse codebook of the 800 most frequent word types by simulated annealing under a bigram language model of the 600 most frequent words, with a noise model that allows single-glyph edits for miscopied words. It splits each text by folio into training, validation and test pages. We enciphered held-out texts of all 16 languages with the two best fits of Cipher-W and Cipher-W2, at their fitted rate of miscopying and with 20, 40 and 60 percent of the words miscopied, and read each text and its page-shuffled version. With all 16 language models, the word solver picks the right language on the validation pages for only 24 of the 64 fitted-rate controls (Table~\ref{tab:wordlang}), so every control is read with its own language, while the manuscript is read with all 16. The word gap is the score per test token of a text minus that of its shuffled version.
	
\begin{figure}[t]
		\centering
		\includegraphics[width=\columnwidth]{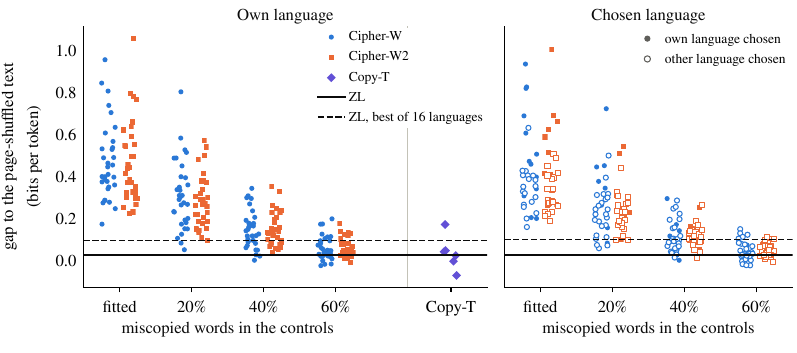}
		\caption{Word gap of the enciphered controls in 16 languages at the fitted rate and at three higher rates of miscopying, and of six texts of Copy-T, against the gap of ZL under the language chosen on its validation pages (solid line) and under the best of 16 languages (dashed line). Each text of Copy-T is read with the language chosen on its own validation pages, as ZL is for the solid line. In the left panel each control is read with its own language. In the right panel, from a separate run, each control and its shuffled version are read with the languages chosen on their own validation pages, and the lines show ZL in that run.}
		\label{fig:decoder}
	\end{figure}

	With the codebook known, the word solver recovers 49\% of the plaintext words of the fitted-rate controls, and with the codebook unknown 12\%. Their word gaps lie between 0.17 and 1.05 (Figure~\ref{fig:decoder}, left). The manuscript reaches 0.026 under the language chosen on its own validation pages and 0.096 under the best of all 16. All 64 fitted-rate controls lie above both values, even when each control and its shuffled version are read with the languages chosen on their own validation pages, as for the manuscript (Figure~\ref{fig:decoder}, right). At 20 percent miscopying 2 of 64 controls come down to the best value of the manuscript, at 40 percent 20 and at 60 percent 50. At 60 percent the controls lie at a distance of at least 0.73 from the manuscript on the 22 fitted statistics, so they are no longer texts that resemble it. Six texts of Copy-T reach word gaps between $-0.070$ and 0.171, around the value of the manuscript.
	
		\begin{table}[t]
		\centering
		\small
				\caption{The 64 controls at each rate of miscopying in the separate run of Figure~\ref{fig:decoder}, read with all 16 language models. Right language counts the controls for which the solver chooses their own language on their validation pages. The median word gap is taken with each control read with its own language and, as for ZL, with the control and its shuffled version read with the languages chosen on their own validation pages. The last two columns count the controls whose gap with the chosen languages is at or below that of ZL in this run, 0.026 under its chosen language and 0.100 under the best of the 16.}
		\label{tab:wordlang}
		\begin{tabular}{@{}lccccc@{}}
			\toprule
			& Right & \multicolumn{2}{c}{Median word gap} & \multicolumn{2}{c}{At or below ZL} \\
			& language & \multicolumn{2}{c}{(bits per token)} & \multicolumn{2}{c}{(of 64)} \\
			\cmidrule(lr){3-4}\cmidrule(l){5-6}
			Miscopied words & (of 64) & Own language & Chosen language & 0.026 & 0.100 \\
			\midrule
			fitted & 24 & 0.436 & 0.385 & 0 & 0 \\
			20\% & 17 & 0.274 & 0.228 & 0 & 6 \\
			40\% & 11 & 0.132 & 0.102 & 3 & 30 \\
			60\% & 5 & 0.054 & 0.036 & 17 & 56 \\
			\bottomrule
		\end{tabular}

	\end{table}
	\paragraph{Abbreviated Latin}
	Cipher-W and Cipher-W2 mostly choose abbreviated Latin as their plaintext (Appendix~\ref{app:priors}), so we also test word codes of it. We split its 48 documents into three folds, keeping documents that share word sequences together. We enciphered a stretch of each fold as long as ZL with the three best parameter sets of each search of Cipher-W and Cipher-W2. Every control is read with a model trained on the other two folds, about 85,000 words. For comparison we did the same with normalised Latin and models of the same size. Abbreviation weakens the word solver. At the fitted rate of miscopying it recovers 5\% of the words of abbreviated Latin with the codebook unknown and 34\% with the codebook known, against 11\% and 44\% for normalised Latin. Of the 36 controls of abbreviated Latin, 13 lie at or below ZL, and 9 of the 36 without miscopying, against 1 of 36 for normalised Latin. Under the three models of abbreviated Latin the mean word gap of six texts of Copy-T lies between 0.05 and 0.14 and that of ZL between 0.05 and 0.10. Read with a 1.5-million-word model of normalised Latin, ZL lies below 32 of the 36 controls but IT below only 20, so the two transliterations disagree. With these models the test cannot decide whether the manuscript is a word code of abbreviated Latin.
	
	\paragraph{A larger model of abbreviated Latin}
	We therefore trained models on 5 million words from 144 Latin manuscripts of CoMMA written between 1200 and 1500, with at most 50,000 words from each \citep{clerice2026comma}. These machine transcriptions keep the abbreviations. We mapped them to the symbols of the control texts, and their share of abbreviation signs is close (10.8\% against 11.3\%). We left out the 21 manuscripts that share word sequences with a document of the abbreviated corpus, and no control text shares more than 0.4\% of its word 5-grams with the text that remains. A model of the 2,000 most frequent words covers 52\% of the words of the control texts, against 41\% for the models of the folds. The word solver reads the 2,500 most frequent word types of a text, and with this model we read the controls of the best parameter set of each search of Cipher-W and Cipher-W2, 12 at each rate of miscopying. ZL then reaches a word gap of 0.028, IT 0.053 and the texts of the best fit of Copy-T in each search 0.053. At the fitted rate, where 4\% of the words are miscopied, 11 of the 12 controls lie above ZL and 10 above IT, with a median gap of 0.111. Without miscopying and at 20 percent the medians are 0.148 and 0.065, and again 11 of 12 controls lie above ZL. Only at 40 percent do the controls come down to the manuscript, with a median of 0.018 and 10 of 12 at or below ZL. A model of the same 5 million words with a vocabulary of 600 gives ZL a gap of $-0.011$, below all 48 controls. The word solver still recovers only 4\% of the words at the fitted rate with the codebook unknown and 41\% with it known, so this test rests on word order. A word code of abbreviated Latin matches the word order of the manuscript only if two words in five are miscopied.

	\paragraph{Herbals and recipes}
	Herbals and recipe books consist largely of lists and set phrases, which could leave little order for the word solver to find. We searched the corpus with a word list of ingredients, plant parts, preparations, doses and illnesses. We enciphered the texts it found and a Latin medical text at the fitted rate and at 20 and 40 percent miscopying. These texts give larger gaps than prose of the same language under the same cipher, by 0.091 on average. None of the 24 texts found by the word list comes down to the level of the manuscript, and one of six texts of the medical work does so at 40 percent miscopying. The genre of the manuscript does not explain its missing word order.
	
	\subsection{Nomenclators}
	\label{sec:nomen}
	
	A nomenclator mixes a word code with a letter cipher, so we read it with both solvers. The letter solver of Section~\ref{sec:reading} takes the code words as units of its language models. For each fit and language, the model keeps whole as many of the most frequent words as the fit codes and spells the other words in letters or Naibbe units. Its controls are held-out texts of 40,000 words, two in each of the 16 languages, enciphered with the best fit of each search. The word solver of Section~\ref{sec:wordcodes} reads controls of the same fits at the fitted rate of miscopying and at 20, 40 and 60 percent.
	
	The tests exclude the nomenclators with one key in many languages. The letter solver reads the controls of the two fits of Cipher-MN in 12 and 15 of the 16 languages and those of Cipher-M in 7 and 13, and in each of these languages ZL and IT lie below both controls. With the word solver, 27 of the 32 fitted-rate controls of Cipher-MN and 21 of 32 of Cipher-M lie above the manuscript in this run (0.026 under its chosen language and 0.100 under the best of the 16). Cipher-M2 escapes most of these tests. The letter solver reads its controls in only two languages in the first search and one in the second, and the word solver places 2 and 6 of its 16 fitted-rate controls above the manuscript. At the fitted rate of miscopying, the word solver recovers at most 8 percent of the code words of any nomenclator. The joint check of Section~\ref{sec:language} separates the first fit of Cipher-M2 from ZL. The second remains open in 9 of the 16 languages (Arabic, Church Slavonic, German, Persian, Middle French, Greek, Hebrew, Italian and Old Occitan).
	
	\paragraph{A solver for the open nomenclator}
	We built a solver for this fit. Its units are the 2,000 most frequent words of a language and the letters of all other words, as in the cipher. Its language model is a bigram model over these units, trained on 14 to 20 million words for Latin, Italian and Middle French and on 84,000 to 1.3 million words for the other 13. Each written word is a homophone of its unit or, at a fixed rate of miscopying, a homophone with one glyph changed, added or dropped. Before it searches for the key, the solver uses expectation maximisation to decide which word types of a text are homophones and which are miscopies of them. It then searches for the key by simulated annealing, at two rates of miscopying and with either one key or one key for each dialect, and keeps the reading that scores best on the validation pages. On these pages one key almost always wins, even on controls made with two keys, so a second version of the solver always uses one key for each dialect. The tests above used only the best parameter set of the fit, and we now use its three best. With each set we enciphered two stretches of held-out text in each of the 16 languages, which gives 96 controls at the fitted rate of miscopying. For the best set we also made controls without miscopying and with 20 and 40 percent miscopying. We also read every control with its true key, which shows how large its gap could be. The gap of a control changes from one reading to the next, with a new seed, by a median of 0.04 and by up to 0.26, so we read every text three times. A control counts as above the manuscript when even its lowest gap in three readings is higher than every gap of ZL in three readings, both under the language chosen on the validation pages of ZL and under the best of the 16 languages. A parameter set is excluded in a language when both of its controls in that language lie above the manuscript under the same solver, and the cipher is excluded when all three sets are. We apply the same rule to the word solver of Section~\ref{sec:wordcodes}.
	
	Under the new solver the gap of ZL is at most 0.147, and at most 0.115 under its second version, which always uses two keys. Both controls of the first parameter set lie above the manuscript in 14 languages, and those of the second set in 15 (Appendix~\ref{app:nomen}). The first set stays open in Arabic, where one control reaches a gap of only 0.074 against 0.115 for ZL, and both sets stay open in Hebrew. The third set is excluded only in Greek, Czech, Old Spanish and Portuguese, so with all three sets the solvers exclude the cipher in these four languages alone. In Middle English, Old French, Latin and Old Norse the tests above had excluded the best set, but the third set stays open. This set writes 8.4 percent of its words as null words, against about 3 percent for the other two, and nearly all of these null words are ol. A third version of the solver allows the most frequent word of each dialect to be a null word and fixes the rate of miscopying at that of the set. Compared with the second version, it raises the lowest gap of 25 of the 32 controls of this set. But the gap of ZL rises as well, to 0.210 in one reading as Old Spanish, and no further language is excluded.
	
	\paragraph{The most frequent word}
	The null words of the third set leave a trace that needs no solver. In its texts ol makes up 8 to 14 percent of the words of each dialect, while the most frequent word of ZL makes up 4.4 percent of Currier A (daiin) and 2.1 percent of Currier B (chedy). We added a check of this share after the solvers had run, but fixed its rule in the code before running it. For every parameter set, language and stretch of held-out text we wrote 50 texts with different seeds, the first being the control itself. A set is excluded in a language when all 100 of its texts in that language give their most frequent word a larger share than ZL does, in Currier A and in Currier B alike. The third set is excluded in all 16 languages, for ZL and for IT, since its 1,600 texts give their most frequent word at least 8.9 percent of Currier A and 8.3 percent of Currier B. The first two sets are not excluded, because in Currier A nearly all of their texts give their most frequent word a smaller share than ZL does.
	
	If a parameter set counts as excluded when either the solvers or this check exclude it, the cipher is excluded in 14 of the 16 languages, for ZL and for IT. We set this combined rule after both tests had run. The cipher remains open in Arabic, where one control of the first set falls short, and the test is undecided in Hebrew. Read with their true key, the Hebrew controls at the fitted rate would reach gaps of at least 0.32, well above ZL, but the solver reads only 46 percent of the letters of the Hebrew controls correctly even without miscopying, against 68 to 98 percent in the other languages.
	
	\subsection{One word, one unit}
	\label{sec:oneunit}
	
	A word code with one plaintext word per Voynich word is not the only way to hide a text in whole words. A scribe could also write one letter per word, choosing the word from several that stand for the letter and inserting null words. Such a code turns the manuscript into a long letter cipher in which each word is a letter, and it hides the grammar of words entirely.
	
	\paragraph{Test}
	For every scheme we learn a table from word types to plaintext units on each half of the text and use it to read the other half. The \textit{gain} is the number of bits per token of a type the table knows that the letter order of a language adds to this reading. A text without plaintext still gains something because the table fits chance patterns, so we set a ceiling for every combination of language and scheme at the largest gain among the texts without plaintext. These are ZL shuffled within pages, texts of the three copying programs of Section~\ref{sec:morecopying} and twin texts, which keep the table and the choices of a code but scramble the order of its plaintext. Codes of one scheme with the same share of nulls and the same way of choosing among homophones form a family, and a family is read when every one of its texts gains more than the ceiling. It is excluded when it is read and neither ZL nor IT gains more than the ceiling, not excluded when it is read and ZL or IT gains more, and undecided when it is not read. A separate test with the mutual information between the starts or ends of neighbouring words excludes codes that write a letter in the first or last two glyphs of each word in all 16 languages and in abbreviated Latin, apart from four of these codes in Hebrew.
	
	\begin{table}[t]
		\centering
		\small
		\setlength{\tabcolsep}{2.5pt}
				\caption{Decisions on codes that write one unit per word, over 16 languages, abbreviated Latin and the families of each scheme. A decision is excluded, undecided or not excluded as defined in the text. Reversed or columnar reading includes boustrophedon, and the groups of pages are the dialects, sections, scribes and quires.}
		\label{tab:oneunit}
		\begin{tabular}{@{}lrrr@{}}
			\toprule
			Codes & Excl. & Undec. & Not excl. \\
			\midrule
			One letter per word, 0--10\% nulls & 85 & 34 & 0 \\
			One letter per word, 25--40\% nulls & 5 & 97 & 0 \\
			Homophones taken in turn & 0 & 17 & 0 \\
			Word spaces written as words & 33 & 1 & 0 \\
			Vowels left out & 1 & 27 & 0 \\
			Nulls at fixed places in the line & 225 & 25 & 22 \\
			First word of a line, own table & 28 & 6 & 0 \\
			Reversed or columnar reading & 93 & 7 & 2 \\
			Letter in the word end & 31 & 3 & 0 \\
			Letters in the parts of a word & 42 & 26 & 0 \\
			Letter in the word start & 0 & 1 & 33 \\
			Two letters per word & 20 & 14 & 0 \\
			One or two letters per word & 3 & 31 & 0 \\
			One syllable per word & 4 & 24 & 0 \\
			Tables by groups of pages & 26 & 80 & 30 \\
			Tables by position in the line & 0 & 36 & 32 \\
			\midrule
			All & 596 & 429 & 119 \\
			\bottomrule
		\end{tabular}

	\end{table}
	
	\paragraph{Results}
	We tested 28 schemes in 16 languages and abbreviated Latin, from one letter per word with up to 40 percent nulls to nulls at fixed places, reversed or columnar reading, letters in parts of words, two letters or a syllable per word and separate tables for dialects, sections, scribes, quires and positions in the line. Table~\ref{tab:oneunit} groups the 1,144 decisions. In 596 the code is read and excluded, in 429 the test is undecided, and in 119 the code is read and the manuscript gains more than the ceiling. The test excludes every code of one letter per word with up to ten percent nulls that it reads, and most codes with nulls at fixed places, reversed or columnar reading, a letter in the end of a word or in its parts, and a separate table for the first word of each line. It leaves undecided the codes with many nulls, without vowels or with homophones taken in turn, and most codes with several letters per word. Hebrew, Arabic and Persian, written as consonants, leave most of their tests undecided.
	
	The manuscript exceeds the ceiling in three kinds of case, always by little. Under codes with a letter in the word start, ZL exceeds the ceiling in 33 of 34 cases by a median of 0.012 bits. A code of this kind gains far more, 0.76 bits in German, where ZL gains 0.054 against a ceiling of 0.048. The manuscript's mutual information between the starts of neighbouring words also lies far below what such a code produces. Under separate tables for groups of pages or for positions in the line, texts without any plaintext already gain a median of 0.28 bits, ZL exceeds them by a median of 0.024, and the codes themselves are read correctly for only one token in ten, so these tests have little power. In the remaining 24 cases, mostly nulls at fixed places in Arabic, Persian and Old Norse, ZL exceeds a ceiling near zero by a median of 0.002 bits. The decoded halves of the manuscript are unreadable in every case. In 392 of the 596 exclusions the ceiling comes from a copying text. Without these texts ZL would often exceed it, since like any text with local order it gains more than its shuffled version and the twins.
	
	\section{Other Ways to Carry Meaning}
	\label{sec:other}
	
	\paragraph{Filler around a real text}
	A scribe could hide a real text among filler words copied from their neighbours. Removing such filler should then reveal the word pairs of the real text. We removed every word that copies or nearly copies one of the two words before it or the word above it and its two neighbours, 16 percent of the text, and compared the word-pair replication with that after removing the same number of random words. ZL rises by 0.034 ($z = 3.1$), but IT only by 0.022 ($z = 1.3$), and texts of Copy-H, which contain no real text, rise by up to 0.09. Plaintexts padded with 15 percent of copied filler rise by 0.012 to 0.071. The rise fails the rule we fixed in advance, which asks for $z \ge 2$ in both transcriptions and a rise beyond that of the copying texts, so it is no evidence of a hidden text.
	
	\paragraph{Halves of hidden words}
	The words of the manuscript could be halves of real words, with each real word spread over two neighbouring written words. We split every word in the middle and rejoined the halves of neighbouring words in three ways. When the halves are rejoined in the way the text was written, natural texts written in such a scheme lose 14 to 54 percent of their distinct units, while plain natural texts gain 19 to 229 percent. Under the simplest scheme, which joins the second half of each word to the first half of the next, ZL gains 2.3 percent, inside the range of the copying texts of Copy-L (0.4 to 14.0 percent), and it behaves the same way under the other two. The scheme does not describe the manuscript.
	
	\paragraph{Notation}
	The text could be a notation such as numbers, a musical scale or counted strokes. Labels that numbered the zodiac figures would recur in every sign, so the labels of two signs would match one to one, and each label would resemble the labels of other signs more than those of its own. They do the opposite ($z = -14.0$ and $z = 8.6$), while numbered controls in the same rings are detected in 90 to 100 percent of cases. In chant written with pitch letters, letters that replace each other in similar words are neighbouring steps of the scale in a share 0.15 to 0.23 above chance. Ordered by how they follow each other within words, glyphs that replace each other in the manuscript show no such preference (0.000). The length of a run of the glyphs i or e depends slightly on the rest of the word. On a scale where a pure count gives 0 and natural spelling gives 0.76 to 0.86, the runs reach 0.11 in ZL and 0.20 in IT, just below and inside the range of the copying texts (0.13 to 0.50), so they do not point to a count.
	
	\paragraph{Rings}
	The scribe could have written the running text by reading its words off the text along the circles of the manuscript, which we call rings. If so, words that stand next to each other in the running text would often stand next to each other on a ring, and some words of the running text could be found on a ring only by reading across one of its word breaks, where the scribe misread a space. ZL has 81 rings of at least four words, with 2,334 words on 27 pages. Among the pairs of neighbouring words in the running text whose two words both stand on one ring, 19.1 percent also stand next to each other on that ring, in the same order. That is more than on rings with their words shuffled (13.5 percent, $p = 0.001$) but no more than on rings made from stretches of the running text (20.0 percent, $p = 0.78$). Of the words of four to twelve glyphs in the running text, 4.1 percent can be found on a ring only by reading across one of its word breaks, against 3.8 and 4.2 percent in the two comparisons. The rings share the word order of the running text no more than any stretch of it does, and show no sign that the scribe read from them.
	
	\paragraph{Labels}
	The 1,159 label words of the manuscript occur in the text of their own page as often as labels moved to random other pages do (0.144 against 0.123 to 0.146, $p = 0.09$), while the labels of Copy-H, which copies them from the page, do so twice as often (0.30 to 0.36). Over half of the real labels start with o, against a fifth of the generated ones. We also paired 22 plants of the herbal section with their counterparts in the pharmaceutical section, where they are drawn again with labels. The label of a pharmaceutical drawing is no more similar to the text of its herbal page than to the text of neighbouring herbal pages (a difference of $-0.12$, $p = 0.99$), and in only one of 22 pairs does the label appear on its page. The labels form a vocabulary of their own, as \citet{schwartz2026voynich} finds, but we see no link between a label and the herbal page of its plant.
	
	\paragraph{Pictures}
	If the text of a herbal page describes its plant, pages with similar texts should show similar plants, and some words should go with features of the drawing. We measured twelve features of the plant on 118 herbal pages from the scans, among them its size, the shares of green, blue and brown paint, the width of its roots and the number and shape of its leaves. We first removed from texts and features what quire, scribe, text length and layout explain. As a reference we shifted the pictures along the book, so that each text was paired with the picture of another page. A link also had to be stronger with the page's own picture than with the pictures one and two pages away, so that a slow drift shared by texts and drawings does not count. We calibrated this rule on random pictures that drift like the real ones. Pages with similar texts do not show more similar plants ($p = 0.94$), and no word predicts a feature after correction for the number of words and features tested (best $p = 0.48$). To learn what the test can find, we planted words. Like a copied word, each planted word also appears on the following page in three cases out of ten. We gave each feature three planted words. When they stand on 80 percent of the pages where the feature lies in its top third and on 10 percent of the other pages, the test finds the link in every trial. At 60 against 10 percent it finds it in 63 percent of the trials, and at 50 against 10 percent in 6 percent. With one planted word per feature it finds no link even at 80 against 10 percent. The text thus has no strong link to the drawings carried by several words per feature, but a weaker link, or one carried by a single word, would escape the test.
	
	\section{Could a Short Message Hide in the Text?}
	\label{sec:short}
	
	Everything above tests a text whose meaning runs through the whole book. A scribe could also have hidden a short message in a text that is otherwise copied, for instance in the first word of every line or in his choice among words he could copy.
	
	\paragraph{Capacity}
	The choices of the fitted Copy-H, where to copy from and which candidate to keep, carry at most 7.4 to 8.3 bits per word, or 35 to 39 kilobytes over a text of the manuscript's length. Plaintexts in our 16 languages need 10.9 to 18.0 bits per word after compression, abbreviated Latin 17.5 and ZL itself 13.6. Choices alone could thus carry a message between two fifths and three quarters as long as a plaintext of the same length, so capacity rules nothing out.
	
	\paragraph{The choices of the scribe}
	A message that steered the scribe's choices would leave structure in the sequence of choices. We parsed the manuscript and the texts of Copy-H into likely copying choices with the same parser. The sequence of choices in the manuscript differs from that of the generator (Mahalanobis distance 26.8, $p = 0.01$, or 5.6 and $p = 0.03$ when the spread of the fitted parameters is included). The test has little power. The parser recovers the true choice of the generator for only 38\% of words, and messages planted in the choices of generated texts are found in only 10 to 50 percent of cases. Since the generator also misses many other statistics, the difference does not point to a message.
	
	\paragraph{Planted messages}
	We plant some to find how long a message must be before our test always finds it. We write messages of 50 to 5,000 letters into seven channels of ZL, the first and the last word of every line, the first word of every page and of every paragraph, the labels, the first line of every paragraph and every word of the text. A message takes up consecutive positions of its channel, every second or third position, or positions spread evenly over the whole channel. Each letter becomes one of the frequent word types of the channel, with one type per letter, three per letter on average or all types of the channel dealt out among the letters. The messages come from 17 plaintexts, the 16 languages and abbreviated Latin, and we plant three of them for every combination of channel, placement, code, plaintext and length. We read every text with the split-half test of Section~\ref{sec:oneunit}, learning the table on one half of the channel's stretches in a window and reading the other half, over windows of the whole book, of sections, of quires and of groups of five or ten pages. For every kind of window we take the largest gain of its windows and standardise it over the references, and a message counts as found when the largest of these standardised gains exceeds that of every reference. The references are ZL itself, twelve versions of ZL shuffled within pages and twins of planted messages no longer than the message. ZL is one of the references, and a planted message could not beat it every time if ZL already held such a message, so the shortest length from which every planted message is found bounds a message of that kind in the manuscript.
	
	Only one kind of message gets a bound in nearly every plaintext, a message with one word type per letter in consecutive positions (Table~\ref{tab:bounds}). Among the first words of paragraphs every such message of 500 letters is found in all 17 plaintexts, and in four of them already from 200 letters. In every plaintext except Hebrew, such messages need 500 to 2,000 letters to be always found among the first or last words of lines, 1,000 to 2,000 in the first lines of paragraphs and 1,000 to 5,000 in the running text. Among the labels they need 200 or 500 letters in 12 plaintexts. The 225 first words of pages are too few for the test, and no kind of message gives a bound there in more than two plaintexts. A message with three word types per letter is always found in consecutive positions only from 2,000 letters on, in 2 to 11 plaintexts depending on the channel. Only the channels with more than 4,000 slots are long enough for that. A message with all types of the channel dealt out among its letters is found there no more often than chance. Messages in every second or third position or spread over the whole channel are always found in at most seven plaintexts. ZL itself stands above all twelve shuffled versions in 12 of the 119 combinations of channel and plaintext, against about 9 expected by chance, and its best readings are scattered letters without words. ZL thus shows no sign of a message, but most ways of writing one escape the test, and the others are found only from these lengths on.
	
	\begin{table*}[t]
		\centering
		\scriptsize
		\setlength{\tabcolsep}{3.5pt}
				\caption{Planted messages that are always found. A cell gives in how many of the 17 plaintexts all planted messages are found from some length on, and in brackets the range of these lengths in letters. A dash means none. Slots are the words of ZL in the channel, and longest is the longest message planted in consecutive positions. A message in every second or third position takes two or three slots per letter, so in some channels these placements stop at shorter lengths. Messages with all types of the channel dealt out among the letters give no bound, and codes with three types per letter give one only where shown and in one plaintext among the first words of pages, at 50 letters in every third position.}
		\label{tab:bounds}
		\begin{tabular}{@{}lrrccccc@{}}
			\toprule
			& & & \multicolumn{4}{c}{One word type per letter} & Three types \\
			\cmidrule(lr){4-7}\cmidrule(l){8-8}
			Channel & Slots & Longest & consecutive & every second & every third & spread & consecutive \\
			\midrule
			First word of a line & 4,355 & 2,000 & 16 (500--2,000) & 5 (1,000--2,000) & -- & 6 (2,000) & 7 (2,000) \\
			Last word of a line & 4,355 & 2,000 & 16 (500--2,000) & 7 (1,000--2,000) & -- & 3 (2,000) & 3 (2,000) \\
			First word of a page & 225 & 150 & -- & 2 (100) & -- & -- & -- \\
			First word of a paragraph & 761 & 500 & 17 (200--500) & 1 (200) & -- & 1 (500) & -- \\
			Labels & 1,159 & 500 & 12 (200--500) & 2 (200--500) & -- & 1 (500) & -- \\
			First line of a paragraph & 7,566 & 2,000 & 16 (1,000--2,000) & 3 (2,000) & -- & -- & 2 (2,000) \\
			Every word & 37,554 & 5,000 & 16 (1,000--5,000) & 5 (5,000) & 1 (5,000) & -- & 11 (2,000--5,000) \\
			\bottomrule
		\end{tabular}

	\end{table*}
	
	\section{Discussion}
	\label{sec:discussion}
	
	\paragraph{What the word statistics say}
	The comparison of programs turns on details. A copier that changes words with a learned model of word shapes loses to a cipher, and the copier that \citet{timm2020generating} built from a close study of the manuscript wins on the statistics that no program was fitted to. Its rules swap whole glyph groups for similar ones, attach and drop prefixes and join the front of one word to the back of another. Applied to almost every copied word, these operations produce the network of similar words, the drift of the vocabulary and the shift in glyph use between the dialects. Ciphers with two keys or with only the word classes of a real text do equally well, and the class cipher does it even with two classes. What these programs share is a scribe who builds many words from words he can see. The word statistics of the manuscript therefore say a great deal about how it was written and little about whether it means anything.
	
	\paragraph{What the trace of language says}
	Word pairs that recur in lines across the book are the most direct sign of language we found, and the manuscript has little of it. Word codes of an ordinary text keep the pattern unless a large share of their words is miscopied, and a heavily miscopied word code no longer resembles the manuscript. The cipher account thus needs one of three things. Its words could stand for letters or other units smaller than words, its plaintext could be a text whose word pairs recur as weakly as those of abbreviated Latin, or it could mix whole words and letters as a nomenclator does. The letter ciphers and the one-unit codes of Section~\ref{sec:decipher} test the first and find no reading where they have power. The second is a word code of abbreviated Latin with one key for each dialect. Its word pairs come close to those of the manuscript, and it predicts the held-out statistics about as well as Copy-T. A large model of abbreviated Latin still finds far more word order in its controls than in the manuscript, unless two words in five are miscopied. That is far more miscopying than the fits choose, so this cipher does not fit the manuscript either. The third is the one cipher our tests leave open, and only in two languages. A nomenclator with two keys matches the word pairs of the manuscript and passes the held-out check, but its scribe miscopies more than half of its words. A solver built for it and a check of how often it writes its most frequent word exclude it in 14 of the 16 languages. It remains open in Arabic and Hebrew (Section~\ref{sec:nomen}).
	
	\paragraph{The boundary between words}
	The last glyph of a word constrains the first glyph of the next more strongly in the manuscript than in our plaintexts and in most of our generated texts, and this link carries most of the structure that a solver finds. \citet{currier1976findings} saw it in the pages of Currier B, \citet{smith2019glyph} measured it across word breaks, and \citet{schwartz2026voynich}, \citet{rozanova2026glyph} and \citet{parisel2026layered} confirm that no generator reproduces it. Its strength comes from the glyphs. Neighbouring glyphs depend on each other twice as strongly as neighbouring letters in our plaintexts, and a space weakens this about as much as in those texts. On the scans the gaps between glyphs come in two sizes, and glyph pairs that often stand on either side of a space are written further apart even inside words. The link does not come from where words stand in the line, and it ends at the line break, where the words of every plaintext stay linked. Programs that add the link through a writing habit, a rule for spaces or strings of glyphs copied across the spaces reach it but place the spaces differently. The strings of glyphs, which never run on into the next line, also end it at the line break, while the rule for spaces carries it into the next line. A word written whole in one place of the manuscript is often split into two words elsewhere, and these programs do so hardly more often than chance. Except in STRING, the last glyphs of neighbouring words depend on each other far more in their texts than in the manuscript (Section~\ref{sec:link}). A copying account must explain why the dependence between neighbouring glyphs crosses the spaces but not the line breaks, as strings of glyphs copied line by line do, and must still place the spaces as the manuscript does. So must any cipher, since one that replaces whole words or letters has no reason to carry the dependence across a space.
	
	\paragraph{Calibration}
	Every negative result in this paper comes with controls that the same test reads. Claims to have read the manuscript can be checked in the same way. Our own earlier attempts proposed word meanings that such checks did not support (Appendix~\ref{app:impl}). A solver that cannot read controls made by the cipher it assumes says nothing about the manuscript, and a solver that reads them but finds nothing in the manuscript is evidence against that cipher. Twin texts, which keep the code and lose the language, show how much a solver finds in a text with no language at all, and this amount is large for codes with many homophones or many tables.
	
	\paragraph{Other undeciphered texts}
	The programs of Section~\ref{sec:programs} are built for the manuscript, but the tests are not. The test of word pairs needs a transcription that marks words and lines, and natural texts laid into the same pages and lines. It suits long texts with word breaks. The short inscriptions of the Indus script, or the signs of Rongorongo, which are not divided into words \citep{tanakaishii2015constancy}, give it little to count. The measures of Section~\ref{sec:published} can be computed on such texts, but like our other tests they need meaningless texts made by a plausible process as controls. The calibrated solvers need only the cipher or code to be tested and texts in the candidate languages, so the check of claimed readings above works for any script. Planted messages show how long a message must be before a test can see it, so a test that finds nothing still limits what could hide in the text.
	
	\section{Conclusion}
	
	We fitted the copying and the cipher accounts of the Voynich manuscript under equal conditions and judged them on statistics outside the fit. A copying process with the rules of Timm and Schinner predicts these statistics better than any cipher with one key, but ciphers with two keys, a nomenclator among them, or with only word classes do equally well, so the word statistics cannot decide between the accounts. Four published measures of language cannot decide either, because copying programs and texts with language both reach the manuscript on all of them. A trace of language that needs no decipherment decides more. The word pairs that share a line in the manuscript recur across the book at half the strength of the weakest of 28 natural texts in 16 languages, a level that word codes reach only with a weak plaintext or with so many miscopied words that they no longer resemble the manuscript. Decipherment calibrated on enciphered controls finds no reading for letter ciphers in 15 languages, all but Hebrew, for word codes in 16 or for 28 ways of writing one unit per word in 17. A word code of abbreviated Latin comes close to the word pairs of the manuscript, but its controls show more word order than the manuscript unless two words in five are miscopied. One cipher remains open, a nomenclator with two keys whose scribe miscopies more than half of its words. It matches the word pairs of the manuscript, but a solver built for it and a check of how often it writes its most frequent word exclude it in 14 of the 16 languages, all but Arabic and Hebrew. The strongest structure left in the text is the link between neighbouring words. It reflects glyphs bound twice as tightly as letters in the languages we tested, and no program reproduces it together with the spaces of the manuscript. It does not come from where words stand in the line, and it ends at each line break, which the link of every plaintext crosses. A program that copies strings of glyphs line by line ends it there as well, but places the spaces differently.
	
	Across nearly all languages our tests rule out only one kind of hidden message, one fixed word per plaintext letter in consecutive positions, and in most channels only from 200 to 5,000 letters on. Messages with several words per letter or with their letters spaced apart could still hide in the manuscript.
	Our negative results all rest on controls that the same tests read. Such controls can be built for any other undeciphered text, and they are what turn a statistic or a failed decipherment into evidence.
	
	\section*{Limitations}
	
	Our conclusions hold for the programs, codes and languages we tested. A cipher with a key that changes the table along the text, transpositions other than reversal and columns, and plaintexts transformed before enciphering are not among them. Our syllables follow a simple rule shared by all languages, and the one-unit tests cover units of at most two letters and the 600 most frequent syllables. The languages are those we could obtain, and the Czech text is from the late sixteenth century. Timm's rules were derived from the manuscript, and the habits of Copy-H take their tables from it, so these programs bring knowledge of the manuscript that the ciphers lack. All three solvers use bigram models, the word solver only the 600 or 2,000 most frequent words. It cannot tell the language of a text reliably and recovers few words of abbreviated Latin even with the larger model, so its test of that code rests on word order. The letter solver cannot read Hebrew controls or most controls of a nomenclator with two keys, and the solver built for that nomenclator cannot read its Hebrew controls either. We tested the open nomenclator only with the three best of the 25 parameter sets of its fit. For short messages the test gives a general bound only for the simplest code, since homophones or gaps between the letters keep planted messages from being found reliably. Both transliterations follow the same conventions for word breaks and glyphs, which affect several statistics, the bond between neighbouring glyphs among them. We read the uncertain spaces of ZL as word breaks, although on the scans three in four of them are narrower than a word space. Our test of the herbal drawings would miss a weak link between a page and its plant, or one carried by a single word per feature. Of the published measures of language we computed four, without the revised threshold that \citet{sproat2014statistical} gives for one measure of \citet{lee2010pictish} and without the block entropies of later work on the Indus script. No program passes the joint check over all 37 statistics, so every account we tested remains incomplete.

	\appendix
	\makeatletter
	\newcommand{\appsection}[1]{\appendixsection{#1}\phantomsection\protected@edef\@currentlabel{\Alph{section}}}
	\makeatother
	\renewcommand{\dbltopfraction}{0.85}
	\renewcommand{\textfraction}{0.1}
	\appsection{Statistics}
	\label{app:stats}
	
	Table~\ref{tab:stats} defines all statistics of Section~\ref{sec:programs} and gives their values for both transliterations. Near means one glyph changed, inserted or deleted. Shuffled texts keep the words of each page and change their order.
	
	\begin{table*}[t]
		\centering
		\scriptsize
		\setlength{\tabcolsep}{4pt}
				\caption{The 41 statistics. Fitted statistics enter the distance $D$, held-out statistics only the checks, and layout statistics are reported only, except in the fits of Sections~\ref{sec:morecopying} and~\ref{sec:link}.}
		\label{tab:stats}
		\begin{tabular}{@{}p{3.4cm}p{7.6cm}rr@{}}
			\toprule
			Statistic & Definition & ZL & IT \\
			\midrule
			\multicolumn{4}{l}{\textit{Fitted statistics (22)}} \\
			Repeat of the previous word & rate of a word equal to the word before it, log ratio to the page-shuffled text & 0.185 & 0.362 \\
			Repeat 2--5 words back & the same for words two to five positions back & 0.281 & 0.183 \\
			Near repeat, previous word & rate of a word one glyph change from the word before it, log ratio to the shuffled text & 0.237 & 0.218 \\
			Near repeat 2--5 back & the same for words two to five positions back & 0.173 & 0.200 \\
			Near repeat 6--20 back & the same for words six to twenty positions back & 0.085 & 0.089 \\
			Similar to the word above & equal or near to the word at the same place one line up, log ratio to a random word of that line & 0.143 & 0.162 \\
			Similar across the page & equal or near pairs at least 20 words apart on one page, log ratio to pairs from different pages & 0.520 & 0.520 \\
			Page-specific vocabulary & share of a page's types found on no other page, log ratio to words shuffled across pages & 0.161 & 0.192 \\
			Types with a near neighbour & share of types with another type one glyph change away & 0.892 & 0.891 \\
			Near neighbours per type & mean number of such types & 14.952 & 15.303 \\
			Neighbours and frequency & Spearman correlation of type frequency and number of near neighbours & 0.587 & 0.577 \\
			Type-token ratio & types per token in the first 10,000 tokens & 0.290 & 0.297 \\
			Hapax rate & types seen once per token in the first 10,000 tokens & 0.199 & 0.205 \\
			Top 10 types & share of tokens in the ten most frequent types & 0.126 & 0.123 \\
			Zipf slope & slope of log frequency on log rank, first 500 ranks & $-$0.930 & $-$0.915 \\
			Word length & mean length in glyphs & 4.461 & 4.526 \\
			Spread of word length & standard deviation over mean of word length & 0.374 & 0.353 \\
			Length and frequency & Spearman correlation of frequency and length, types seen twice or more & $-$0.245 & $-$0.257 \\
			Adjacent-word information & mutual information of adjacent words (200 top types), minus the page-shuffled value & 0.116 & 0.099 \\
			Glyph entropy & glyph entropy over the log of the glyph inventory & 0.771 & 0.794 \\
			Glyph predictability & entropy of a glyph given the previous one, over the glyph entropy & 0.585 & 0.586 \\
			Glyph position rigidity & how strictly glyphs keep to one place in a word & 0.622 & 0.633 \\
			\midrule
			\multicolumn{4}{l}{\textit{Held-out statistics (15)}} \\
			Vocabulary growth & slope of log types on log tokens at 2,000 to 32,000 tokens & 0.749 & 0.752 \\
			Types that occur twice & share of types seen exactly twice & 0.112 & 0.111 \\
			Top 100 types & share of tokens in the 100 most frequent types & 0.441 & 0.430 \\
			Distance back to a repeat & log of the median distance on the page to the last occurrence of a word & 1.477 & 1.462 \\
			Near neighbour among last 10 & share of words with a near (not equal) word among the ten before it & 0.270 & 0.266 \\
			Repeat within a line & share of words already written in the same line & 0.040 & 0.038 \\
			Reused word pairs & share of adjacent pairs in a line seen earlier in the text & 0.165 & 0.153 \\
			Word entropy given previous & entropy of a word given the previous word, over the word entropy & 0.432 & 0.423 \\
			Overlap of adjacent pages & vocabulary Jaccard of consecutive pages, log ratio to pages ten or more apart & 0.508 & 0.529 \\
			Short words & share of words of at most two glyphs & 0.121 & 0.101 \\
			Long words & share of words of at least seven glyphs & 0.108 & 0.108 \\
			Shared contexts of near neighbours & context cosine of type pairs one glyph apart over that of all pairs (200 top types) & 1.652 & 1.682 \\
			Vocabulary drift & slope of log vocabulary Jaccard of two pages on log of their distance (1 to 128 pages) & $-$0.121 & $-$0.129 \\
			Currier A/B word divergence & Jensen--Shannon divergence of word frequencies, Currier A against B pages (bits) & 0.467 & 0.477 \\
			Currier A/B glyph divergence & the same for glyph pairs inside words, word edges included & 0.093 & 0.093 \\
			\midrule
			\multicolumn{4}{l}{\textit{Layout statistics, reported only (4)}} \\
			Line-initial glyphs & divergence of first glyphs, line-initial against other words, minus the shuffled value & 0.185 & 0.185 \\
			Line-final glyphs & the same for last glyphs of line-final words & 0.080 & 0.081 \\
			Paragraph-initial glyphs & first glyphs of paragraph-initial against other line-initial words & 0.475 & 0.460 \\
			Line-initial length & standardised length difference of line-initial words & 0.373 & 0.340 \\
			\bottomrule
		\end{tabular}

	\end{table*}
	\paragraph{Coupling and copying choices}
	The fits of Sections~\ref{sec:morecopying} and~\ref{sec:link} add twelve statistics. The six measures of coupling are mutual information in bits. Five of them pair neighbouring words in a line, by the last glyph of one and the first glyph of the next, by their first glyphs, by their last glyphs, by the first glyph of one and the last glyph of the next, and by the first and last glyph of each word together. The sixth pairs the last glyph of a line with the first glyph of the next line. The six shares of copying choices come from a parser that gives each word the most direct source a copying scribe could have used, a word equal to it or one glyph away. They are the shares of words that continue the previous copy, that copy a word one to ten words back on the page, that copy one more than ten words back, that copy the word above or a word of the previous page, counted together, and that are new, and the share of copies that are exact.
	
	\paragraph{Published measures of language}
	Section~\ref{sec:published} computes four measures on the whole text in the layout of ZL, the text along circles and radii included. For the classifier of \citet{lee2010pictish} we follow their Equations 2.3 and 2.4 and their Figure 6. They insert a start and end mark at every comma and full stop and ignore spaces. We mark the start of the text and the end of every line, and the glyphs of neighbouring words follow each other directly. The classifier first takes the number of different pairs of neighbouring units per different unit and adds seven times the number of pairs seen once divided by the number of all pairs ($C_r$). A text below 4.89 counts as repetitive, like heraldic signs and code characters. For other texts the classifier divides the entropy of a unit given the one before it by the logarithm of the number of different pairs per different unit ($U_r$). It calls the text letters below 1.09, syllables from 1.09 to 1.37 and words above. For the repetition measure of \citet{sproat2014statistical} we follow the account of \citet{raghavendra2026nonspecificity}. A unit counts as a repeat if the same unit occurred earlier in its word, line or paragraph, and as an adjacent repeat if it equals the unit just before it. The measure divides the number of adjacent repeats by the number of repeats, both summed over the text. Within a line, the last glyph of a word and the first glyph of the next are neighbours. The conditional entropy of \citet{rao2009entropic} keeps the most frequent units and counts pairs of neighbouring units that are both kept. A unit that is not kept leaves a gap, so the units on its two sides do not form a pair. Glyph pairs are counted inside words, or inside lines with one extra unit for each space. As \citet{rao2009entropic} did, we estimate the probability of a unit given the one before it with interpolated modified Kneser--Ney smoothing \citep{chen1999empirical}. Each first unit is weighted by its share of the pairs. Yule's K is 10,000 times the difference between the sum of the squared frequencies of all types and the number of tokens, divided by the square of the number of tokens. The 66 values are eight of the classifier (its two ratios, the entropy it divides and its class, for glyphs and for words), eight of repetition (its share and class in four settings), 48 of the conditional entropy (24 settings, as computed and divided by the logarithm of the number of units kept) and two of Yule's K (for glyphs and for words). For a class, a group reaches the manuscript if at least half of its texts fall in the class of the manuscript.
	\appsection{Priors and Fitted Parameters}
	\label{app:priors}
	
	Table~\ref{tab:priors} lists the prior of every parameter and its median among the 100 best simulations of each program in the first and second search. Probabilities are per word unless noted. Copy-W, Copy-T, Copy-N and Cipher-N were fitted with the narrower priors we set first and the other ciphers with wider priors after their first fits ran into the bounds. The median sharpening of W still lies near its upper bound in four cipher fits, but the best fits of the second searches of Cipher-W, Cipher-W2 and Cipher-S lie at 16.5 to 19.4 and come about as close to the manuscript as the first searches (Table~\ref{tab:results}). The most frequent plaintext among the 100 best simulations is the abbreviated Latin for Cipher-W (97\% and 100\%) and Cipher-W2 (56\% and 70\%), Middle French with one document per page for Cipher-S (81\% and 60\%) and Cipher-C (49\% and 22\%), Italian for Cipher-N (30\% and 34\%), Middle French for Cipher-M (30\% and 25\%), Middle English in the first search and Middle French in the second for Cipher-M2 (30\% and 35\%), and Italian with one document per page for Cipher-MN (37\% and 58\%). The median number of code words of Cipher-M and Cipher-M2 lies near its upper bound of 2,000 in three of their four fits.
	\begin{table*}[t]
		\centering
		\caption{Priors and fitted medians of all programs (first / second search). Cipher-N used the narrower priors of the first round, among them 0 to 0.3 for miscopied words and 1 to 400 for the look distance. In Cipher-MN the homophone parameters apply to the code words only, and the other words are spelled with the Naibbe tables.}
		\label{tab:priors}
		\scriptsize
		\setlength{\tabcolsep}{2pt}
		\resizebox{\textwidth}{!}{%
		\begin{tabular}{@{}llccccc@{}}
				\toprule
				Parameter & Prior & \multicolumn{5}{c}{Median among the 100 best simulations, first / second search} \\
				\midrule
				\textit{Copying programs} & & Copy-W & Copy-T & Copy-N & & \\
				\cmidrule(lr){3-5}
				New word & log-uniform 0.002--0.999 & 0.006 / 0.060 & 0.008 / 0.010 & 0.89 / 0.96 &  &  \\
				Copy the word above & uniform 0--1 & 0.021 / 0.10 & 0.21 / 0.062 & 0.013 / 0.039 &  &  \\
				Copy from anywhere on the page & uniform 0--1 & 0.99 / 0.94 & 0.92 / 0.89 & 0.98 / 0.98 &  &  \\
				Mean copy distance (words) & log-uniform 1--2000 & 349 / 57.5 & 39.8 / 56.2 & 12.5 / 33.7 &  &  \\
				Exact copy & uniform 0--1 & 0.10 / 0.12 & 0.023 / 0.008 & 0.21 / 0.17 &  &  \\
				Further W edit & uniform 0--0.99 & 0.11 / 0.17 &  & 0.22 / 0.46 &  &  \\
				Previous page on a first line & uniform 0--1 & 0.050 / 0.087 & 0.63 / 0.75 & 0.17 / 0.17 &  &  \\
				Continue with the next source word & uniform 0--0.99 & 0.15 / 0.11 & 0.96 / 0.87 & 0.64 / 0.91 &  &  \\
				Timm, add or remove a prefix & uniform 0--1 &  & 0.38 / 0.22 & &  &  \\
				Timm, join or split & uniform 0--1 &  & 0.051 / 0.22 & &  &  \\
				Timm, second word & uniform 0--0.5 &  & 0.027 / 0.020 & &  &  \\
				W tilt towards short words & uniform $-$6--1.5 & $-$5.51 / $-$5.01 & $-$0.91 / $-$0.91 & &  &  \\
				W sharpening exponent & uniform 0.3--12 & 11.5 / 11.0 & 8.67 / 8.95 & &  &  \\
				\midrule
				\textit{Ciphers} & & Cipher-W & Cipher-W2 & Cipher-S & Cipher-C & Cipher-N \\
				\cmidrule(lr){3-7}
				Copy bias & uniform 0--1 & 0.99 / 0.65 & 0.44 / 0.63 & 0.93 / 0.45 & 0.53 / 0.91 & 0.99 / 0.99 \\
				Mean look distance (words) & log-uniform 1--2000 & 1052 / 11.7 & 237 / 327 & 12.0 / 2.06 & 10.2 / 229 & 4.31 / 3.33 \\
				Look at the word above & uniform 0--1 & 0.74 / 0.83 & 0.75 / 0.74 & 0.83 / 0.92 & 0.21 / 0.14 & 0.011 / 0.048 \\
				Look anywhere on the page & uniform 0--1 & 0.18 / 0.49 & 0.53 / 0.58 & 0.038 / 0.18 & 0.81 / 0.55 & 0.84 / 0.85 \\
				Look at the previous page & uniform 0--1 & 0.75 / 0.68 & 0.48 / 0.39 & 0.82 / 0.52 & 0.37 / 0.57 & 0.039 / 0.13 \\
				Continue with the next source word & uniform 0--0.99 & 0.007 / 0.73 & 0.74 / 0.90 & 0.31 / 0.35 & 0.47 / 0.42 & 0.63 / 0.87 \\
				Null word & uniform 0--0.4 & 0.014 / 0.008 & 0.056 / 0.068 & 0.008 / 0.087 & 0.037 / 0.067 & 0.003 / 0.005 \\
				Miscopied word & uniform 0--0.6 & 0.080 / 0.053 & 0.11 / 0.13 & 0.17 / 0.20 & 0.52 / 0.55 & 0.062 / 0.091 \\
				Word written twice & uniform 0--0.08 & 0.000 / 0.001 & 0.003 / 0.003 & 0.001 / 0.002 & 0.007 / 0.005 & 0.002 / 0.002 \\
				Variant homophone & uniform 0--1 & 0.42 / 0.18 & 0.69 / 0.45 & 0.41 / 0.27 & 0.55 / 0.32 &  \\
				Homophone factor $h$ & log-uniform 0.001--30 & 0.002 / 0.002 & 0.002 / 0.008 & 0.002 / 0.004 & 0.060 / 0.20 &  \\
				Homophone exponent $\alpha$ & uniform 0--2 & 0.085 / 0.076 & 0.14 / 0.095 & 0.16 / 0.044 & 0.14 / 0.26 &  \\
				Typical forms first $\gamma$ & uniform 0--1 & 0.92 / 0.97 & 0.84 / 0.87 & 0.95 / 0.85 & 0.12 / 0.21 &  \\
				W tilt towards short words & uniform $-$6--4 & $-$2.78 / $-$2.83 & 0.60 / $-$0.76 & $-$2.91 / $-$5.78 & $-$3.70 / $-$5.34 &  \\
				W sharpening exponent & uniform 0.3--24 & 23.9 / 23.4 & 23.2 / 18.6 & 23.6 / 19.5 & 17.0 / 21.3 &  \\
				Units shared by the two keys & uniform 0--1 &  & 0.36 / 0.91 & 0.90 / 0.78 & 0.64 / 0.82 &  \\
				W tilt, Currier B key & uniform $-$6--4 &  & $-$2.61 / $-$2.39 & $-$4.87 / $-$3.03 & $-$3.79 / $-$4.52 &  \\
				W sharpening, Currier B key & uniform 0.3--24 &  & 21.1 / 22.0 & 21.3 / 22.0 & 16.8 / 14.5 &  \\
				Word classes & log-uniform 2--1000 &  &  &  & 940 / 680 &  \\
				\midrule
				\textit{Nomenclators} & & Cipher-M & Cipher-M2 & Cipher-MN & & \\
				\cmidrule(lr){3-5}
				Copy bias & uniform 0--1 & 0.63 / 0.87 & 0.86 / 0.75 & 0.99 / 0.97 &  &  \\
				Mean look distance (words) & log-uniform 1--2000 & 24.7 / 5.69 & 5.42 / 17.1 & 3.55 / 12.1 &  &  \\
				Look at the word above & uniform 0--1 & 0.50 / 0.41 & 0.42 / 0.60 & 0.075 / 0.025 &  &  \\
				Look anywhere on the page & uniform 0--1 & 0.39 / 0.65 & 0.60 / 0.27 & 0.88 / 0.73 &  &  \\
				Look at the previous page & uniform 0--1 & 0.25 / 0.24 & 0.16 / 0.74 & 0.76 / 0.19 &  &  \\
				Look after the last deciding word & uniform 0--0.99 & 0.71 / 0.32 & 0.72 / 0.68 & 0.38 / 0.78 &  &  \\
				Null word & uniform 0--0.4 & 0.083 / 0.077 & 0.12 / 0.10 & 0.002 / 0.009 &  &  \\
				Miscopied word & uniform 0--0.6 & 0.58 / 0.52 & 0.40 / 0.55 & 0.027 / 0.045 &  &  \\
				Word written twice & uniform 0--0.08 & 0.003 / 0.009 & 0.006 / 0.005 & 0.002 / 0.001 &  &  \\
				Variant homophone & uniform 0--1 & 0.39 / 0.37 & 0.094 / 0.31 & 0.54 / 0.93 &  &  \\
				Homophone factor $h$ & log-uniform 0.001--30 & 0.023 / 0.052 & 4.93 / 0.009 & 15.6 / 1.05 &  &  \\
				Homophone exponent $\alpha$ & uniform 0--2 & 0.17 / 0.20 & 0.056 / 0.39 & 1.65 / 1.54 &  &  \\
				Typical forms first $\gamma$ & uniform 0--1 & 0.75 / 0.58 & 0.57 / 0.42 & 0.77 / 0.84 &  &  \\
				W tilt towards short words & uniform $-$6--4 & $-$5.38 / $-$3.37 & $-$5.13 / $-$2.68 & $-$4.33 / $-$2.32 &  &  \\
				W sharpening exponent & uniform 0.3--24 & 17.7 / 20.2 & 14.5 / 17.3 & 20.3 / 10.3 &  &  \\
				Units shared by the two keys & uniform 0--1 &  & 0.55 / 0.43 & &  &  \\
				W tilt, Currier B key & uniform $-$6--4 &  & $-$3.60 / $-$5.09 & &  &  \\
				W sharpening, Currier B key & uniform 0.3--24 &  & 15.5 / 18.4 & &  &  \\
				Code words & log-uniform 5--2000 & 1833 / 1729 & 205 / 1862 & 6 / 5 &  &  \\
				\bottomrule
		\end{tabular}
		}
	\end{table*}
	
	\appsection{All Predictive Checks}
	\label{app:z}
	
	Tables~\ref{tab:z} and \ref{tab:z2} give the $z$-score of ZL under the predictive sample of every program and search. Bold values lie outside the central 95\% of the sample. For IT at most three statistics fall on the other side of the central 95\% than for ZL, the Mahalanobis distance changes by at most 1.6, and the order of the programs stays the same.
	
	\begin{table*}[t]
		\centering
		\scriptsize
		\setlength{\tabcolsep}{3pt}
				\caption{$z$-scores of ZL under the predictive samples of the copying programs, Cipher-W, Cipher-N and Cipher-W2. Bold marks values outside the central 95\% of the sample.}
		\label{tab:z}
		\begin{tabular*}{\textwidth}{@{\extracolsep{\fill}}lrrrrrrrrrrrr@{}}
			\toprule
			& \multicolumn{2}{c}{Copy-W} & \multicolumn{2}{c}{Copy-T} & \multicolumn{2}{c}{Copy-N} & \multicolumn{2}{c}{Cipher-W} & \multicolumn{2}{c}{Cipher-N} & \multicolumn{2}{c}{Cipher-W2} \\
			\cmidrule(lr){2-3}\cmidrule(lr){4-5}\cmidrule(lr){6-7}\cmidrule(lr){8-9}\cmidrule(lr){10-11}\cmidrule(lr){12-13}
			Statistic & 1st & 2nd & 1st & 2nd & 1st & 2nd & 1st & 2nd & 1st & 2nd & 1st & 2nd \\
			\midrule
			\multicolumn{13}{@{}l}{\textit{Fitted statistics}} \\
			Repeat of the previous word & $-$0.9 & $-$0.7 & $-$0.7 & $-$1.1 & $-$0.6 & $-$0.6 & 0.1 & $-$0.4 & 0.3 & 0.3 & $-$0.6 & 0.0 \\
			Repeat 2--5 words back & 0.8 & 1.2 & 0.8 & 0.4 & 0.4 & 0.2 & 0.5 & 0.2 & \textbf{1.8} & 1.6 & 1.4 & 1.4 \\
			Near repeat, previous word & 0.2 & 0.6 & $-$0.3 & 0.1 & 0.7 & 0.6 & \textbf{3.2} & 1.8 & 0.4 & 0.0 & 1.1 & \textbf{2.2} \\
			Near repeat 2--5 back & 1.0 & 0.9 & 1.2 & 0.6 & 0.2 & 0.1 & \textbf{2.1} & 2.1 & $-$0.3 & $-$0.5 & 1.2 & \textbf{2.0} \\
			Near repeat 6--20 back & 1.0 & 0.2 & 0.3 & 0.3 & 0.5 & 0.1 & 1.6 & \textbf{2.4} & 0.4 & 0.9 & 1.7 & 1.6 \\
			Similar to the word above & $-$0.2 & $-$0.8 & $-$0.4 & $-$0.1 & 0.1 & $-$0.2 & 1.3 & 1.1 & 0.3 & 0.3 & 0.3 & 0.7 \\
			Similar across the page & $-$0.7 & 0.0 & $-$0.7 & $-$0.8 & \textbf{2.0} & 1.8 & 1.5 & \textbf{2.1} & \textbf{2.0} & \textbf{2.0} & 1.3 & 1.0 \\
			Page-specific vocabulary & $-$0.7 & $-$0.7 & $-$1.9 & \textbf{$-$2.4} & 0.2 & 0.1 & \textbf{6.4} & \textbf{5.9} & \textbf{2.7} & \textbf{2.5} & 0.8 & 1.2 \\
			Types with a near neighbour & $-$1.3 & $-$0.4 & $-$1.6 & \textbf{$-$1.9} & $-$1.1 & $-$1.5 & $-$0.3 & $-$0.6 & \textbf{$-$35.2} & \textbf{$-$28.2} & $-$0.3 & $-$0.3 \\
			Near neighbours per type & \textbf{$-$2.8} & \textbf{$-$2.1} & $-$0.7 & $-$0.3 & \textbf{$-$2.4} & \textbf{$-$2.4} & \textbf{$-$2.8} & \textbf{$-$3.3} & \textbf{$-$4.8} & \textbf{$-$4.0} & \textbf{$-$2.1} & \textbf{$-$2.8} \\
			Neighbours and frequency & \textbf{2.7} & \textbf{2.2} & 0.0 & 0.6 & \textbf{3.6} & \textbf{3.8} & 1.0 & 1.2 & \textbf{5.2} & \textbf{4.1} & \textbf{1.7} & \textbf{1.7} \\
			Type-token ratio & \textbf{$-$3.5} & $-$1.5 & 1.0 & 0.4 & \textbf{$-$3.5} & \textbf{$-$2.4} & \textbf{$-$3.3} & \textbf{$-$4.4} & $-$0.7 & $-$1.2 & $-$1.4 & $-$2.5 \\
			Hapax rate & \textbf{$-$2.8} & $-$1.2 & 1.3 & 0.6 & $-$1.9 & $-$1.3 & \textbf{$-$3.4} & \textbf{$-$4.4} & 0.6 & 0.0 & $-$1.2 & $-$2.1 \\
			Top 10 types & $-$0.2 & $-$0.8 & $-$1.6 & $-$1.5 & 0.5 & 0.5 & $-$0.6 & $-$0.7 & $-$0.9 & $-$0.4 & $-$0.9 & $-$0.6 \\
			Zipf slope & \textbf{$-$2.3} & $-$1.2 & 0.4 & $-$0.2 & 0.9 & 0.9 & \textbf{$-$1.7} & \textbf{$-$2.3} & \textbf{2.6} & \textbf{1.7} & $-$1.5 & \textbf{$-$1.8} \\
			Word length & $-$1.8 & $-$0.6 & 0.3 & 0.1 & \textbf{$-$14.1} & \textbf{$-$6.1} & $-$1.1 & $-$1.1 & $-$1.6 & $-$1.1 & $-$0.2 & $-$0.6 \\
			Spread of word length & 0.0 & $-$0.2 & 1.1 & 0.6 & \textbf{4.5} & \textbf{2.8} & $-$1.0 & $-$0.9 & \textbf{4.1} & \textbf{3.6} & $-$1.5 & $-$1.0 \\
			Length and frequency & \textbf{2.0} & 1.6 & 1.2 & 1.3 & \textbf{$-$2.4} & \textbf{$-$2.3} & 0.5 & 0.7 & 0.0 & $-$0.7 & 0.4 & 0.5 \\
			Adjacent-word information & \textbf{4.0} & \textbf{5.1} & 0.0 & 0.5 & \textbf{11.7} & \textbf{9.2} & 0.9 & 0.7 & \textbf{2.5} & 1.5 & 0.4 & 1.1 \\
			Glyph entropy & 1.4 & 1.1 & $-$0.2 & $-$0.4 & \textbf{$-$3.4} & \textbf{$-$3.4} & \textbf{$-$2.5} & \textbf{$-$3.1} & \textbf{$-$1.1} & \textbf{$-$0.9} & $-$1.7 & \textbf{$-$2.8} \\
			Glyph predictability & \textbf{$-$6.8} & \textbf{$-$3.3} & $-$0.9 & $-$1.8 & \textbf{$-$2.4} & $-$2.2 & \textbf{$-$4.0} & \textbf{$-$4.6} & $-$0.5 & $-$1.0 & \textbf{$-$3.4} & \textbf{$-$4.0} \\
			Glyph position rigidity & \textbf{2.2} & 1.4 & 0.7 & 1.7 & 0.9 & 1.0 & 1.3 & 1.2 & 0.1 & 0.5 & \textbf{1.8} & \textbf{2.2} \\
			\midrule
			\multicolumn{13}{@{}l}{\textit{Held-out statistics}} \\
			Vocabulary growth & \textbf{$-$5.1} & \textbf{$-$2.8} & 0.3 & $-$0.4 & 0.4 & 0.5 & \textbf{$-$5.4} & \textbf{$-$6.9} & \textbf{4.4} & \textbf{2.9} & $-$0.9 & $-$1.4 \\
			Types that occur twice & $-$0.9 & $-$0.8 & \textbf{$-$4.1} & \textbf{$-$4.9} & \textbf{$-$3.1} & \textbf{$-$3.2} & $-$2.0 & \textbf{$-$2.7} & \textbf{$-$6.9} & \textbf{$-$5.1} & $-$1.4 & $-$1.4 \\
			Top 100 types & 3.0 & 1.3 & $-$0.8 & $-$0.1 & \textbf{2.6} & \textbf{2.3} & \textbf{3.8} & \textbf{7.2} & $-$0.7 & 0.0 & 1.9 & 2.2 \\
			Distance back to a repeat & 0.6 & 0.8 & \textbf{1.9} & \textbf{1.8} & \textbf{$-$2.9} & $-$2.3 & 2.1 & \textbf{3.2} & $-$1.6 & \textbf{$-$2.4} & 1.7 & 1.1 \\
			Near neighbour among last 10 & \textbf{$-$2.7} & \textbf{$-$2.0} & \textbf{$-$2.6} & \textbf{$-$2.0} & \textbf{6.9} & \textbf{5.8} & \textbf{4.3} & \textbf{5.5} & 1.4 & 1.3 & \textbf{2.3} & \textbf{4.0} \\
			Repeat within a line & $-$0.5 & $-$0.7 & \textbf{$-$2.2} & \textbf{$-$2.2} & 2.3 & 2.1 & 0.0 & $-$0.4 & 1.9 & \textbf{2.8} & $-$0.9 & $-$0.4 \\
			Reused word pairs & 2.5 & 0.7 & $-$1.4 & $-$0.9 & \textbf{3.5} & \textbf{2.9} & \textbf{2.4} & \textbf{3.6} & \textbf{2.0} & \textbf{2.1} & 0.4 & 1.0 \\
			Word entropy given previous & \textbf{4.0} & 1.7 & $-$0.6 & 0.0 & \textbf{3.3} & \textbf{2.8} & \textbf{5.6} & \textbf{11.2} & $-$0.5 & 0.1 & 1.5 & 2.5 \\
			Overlap of adjacent pages & 2.0 & 2.3 & $-$0.6 & $-$0.8 & \textbf{7.0} & \textbf{10.9} & $-$2.0 & \textbf{$-$4.9} & \textbf{10.2} & \textbf{14.0} & $-$1.3 & $-$1.8 \\
			Short words & 0.6 & 0.0 & 0.4 & 0.0 & \textbf{11.1} & \textbf{4.9} & 0.6 & 0.5 & \textbf{3.5} & \textbf{3.0} & $-$0.7 & $-$0.3 \\
			Long words & $-$1.6 & $-$0.8 & 0.9 & 0.4 & \textbf{$-$6.4} & \textbf{$-$4.1} & $-$1.4 & $-$1.4 & 2.1 & 0.4 & $-$0.5 & $-$0.8 \\
			Shared contexts of near neighbours & $-$0.2 & 0.2 & $-$1.0 & $-$1.0 & \textbf{10.9} & \textbf{8.4} & \textbf{5.5} & \textbf{6.4} & \textbf{6.2} & \textbf{4.9} & 1.2 & 2.2 \\
			Vocabulary drift & \textbf{$-$3.6} & \textbf{$-$4.2} & 0.3 & 0.5 & \textbf{$-$13.5} & \textbf{$-$20.5} & 1.9 & \textbf{5.0} & \textbf{$-$15.2} & \textbf{$-$20.0} & 1.1 & 1.4 \\
			Currier A/B word divergence & $-$0.2 & 0.6 & \textbf{2.1} & 1.4 & \textbf{5.1} & \textbf{5.2} & 0.4 & 0.9 & \textbf{16.8} & \textbf{12.2} & $-$0.4 & $-$1.2 \\
			Currier A/B glyph divergence & \textbf{14.4} & \textbf{15.5} & \textbf{2.7} & \textbf{3.0} & \textbf{107} & \textbf{76.9} & \textbf{101} & \textbf{147} & \textbf{219} & \textbf{188} & \textbf{5.2} & \textbf{3.3} \\
			\midrule
			\multicolumn{13}{@{}l}{\textit{Layout statistics (reported only)}} \\
			Line-initial glyphs & \textbf{204} & \textbf{209} & \textbf{21.7} & \textbf{11.7} & \textbf{343} & \textbf{360} & \textbf{394} & \textbf{352} & \textbf{420} & \textbf{444} & \textbf{383} & \textbf{286} \\
			Line-final glyphs & \textbf{144} & \textbf{148} & \textbf{196} & \textbf{148} & \textbf{162} & \textbf{165} & \textbf{152} & \textbf{154} & \textbf{213} & \textbf{185} & \textbf{142} & \textbf{134} \\
			Paragraph-initial glyphs & \textbf{30.5} & \textbf{25.7} & 0.7 & 1.2 & \textbf{152} & \textbf{173} & \textbf{156} & \textbf{165} & \textbf{171} & \textbf{167} & \textbf{151} & \textbf{164} \\
			Line-initial length & \textbf{19.3} & \textbf{18.1} & \textbf{3.1} & \textbf{3.9} & \textbf{16.0} & \textbf{16.8} & \textbf{15.6} & \textbf{16.9} & \textbf{16.1} & \textbf{17.2} & \textbf{18.4} & \textbf{17.3} \\
			\bottomrule
		\end{tabular*}

	\end{table*}
	
	\begin{table*}[t]
		\centering
		\scriptsize
		\setlength{\tabcolsep}{3pt}
				\caption{$z$-scores of ZL under the predictive samples of Cipher-S, Cipher-C and the nomenclators. Bold marks values outside the central 95\% of the sample.}
		\label{tab:z2}
		\begin{tabular*}{\textwidth}{@{\extracolsep{\fill}}lrrrrrrrrrr@{}}
			\toprule
			& \multicolumn{2}{c}{Cipher-S} & \multicolumn{2}{c}{Cipher-C} & \multicolumn{2}{c}{Cipher-M} & \multicolumn{2}{c}{Cipher-M2} & \multicolumn{2}{c}{Cipher-MN} \\
			\cmidrule(lr){2-3}\cmidrule(lr){4-5}\cmidrule(lr){6-7}\cmidrule(lr){8-9}\cmidrule(lr){10-11}
			Statistic & 1st & 2nd & 1st & 2nd & 1st & 2nd & 1st & 2nd & 1st & 2nd \\
			\midrule
			\multicolumn{11}{@{}l}{\textit{Fitted statistics}} \\
			Repeat of the previous word & 1.0 & 0.4 & $-$0.6 & $-$0.6 & $-$0.7 & $-$0.7 & $-$0.3 & $-$0.3 & 0.5 & 0.4 \\
			Repeat 2--5 words back & \textbf{3.9} & \textbf{4.1} & 1.2 & 1.1 & 2.5 & 1.3 & \textbf{2.5} & \textbf{2.9} & 1.6 & 1.4 \\
			Near repeat, previous word & 1.5 & 1.9 & 1.0 & 0.6 & 0.2 & 0.7 & 0.4 & 1.1 & 0.2 & 0.6 \\
			Near repeat 2--5 back & 3.4 & 2.2 & 0.9 & 0.7 & 0.8 & 0.7 & 0.8 & 1.1 & $-$0.2 & $-$0.2 \\
			Near repeat 6--20 back & \textbf{4.2} & \textbf{2.8} & 0.1 & 0.4 & 0.5 & 0.9 & 0.8 & 1.1 & 0.0 & $-$0.4 \\
			Similar to the word above & 0.9 & 1.6 & $-$0.4 & $-$0.1 & $-$0.3 & $-$0.2 & $-$0.2 & $-$0.1 & $-$0.2 & 0.0 \\
			Similar across the page & \textbf{4.8} & \textbf{5.3} & 0.2 & 0.6 & \textbf{3.6} & 1.7 & 0.7 & 1.4 & \textbf{2.0} & \textbf{1.8} \\
			Page-specific vocabulary & \textbf{4.3} & 1.8 & $-$0.1 & 0.1 & 0.7 & 0.9 & 0.2 & 1.2 & \textbf{2.4} & \textbf{2.2} \\
			Types with a near neighbour & $-$1.2 & $-$0.6 & $-$1.2 & $-$1.1 & $-$0.3 & $-$0.4 & $-$1.3 & $-$0.9 & $-$1.3 & $-$0.8 \\
			Near neighbours per type & $-$2.2 & $-$1.5 & \textbf{$-$2.2} & $-$1.6 & \textbf{$-$1.6} & $-$1.0 & $-$2.0 & $-$1.7 & $-$2.0 & $-$1.0 \\
			Neighbours and frequency & 0.9 & 1.0 & \textbf{1.4} & \textbf{1.6} & 1.0 & 1.6 & \textbf{1.6} & 1.6 & \textbf{2.9} & \textbf{4.1} \\
			Type-token ratio & 0.6 & 0.7 & 0.0 & $-$0.5 & $-$0.2 & 0.0 & 0.3 & $-$0.3 & $-$0.7 & $-$1.4 \\
			Hapax rate & 0.9 & 0.9 & 0.2 & $-$0.4 & $-$0.1 & 0.0 & 0.3 & $-$0.2 & 0.3 & $-$0.1 \\
			Top 10 types & $-$0.6 & $-$1.1 & $-$1.3 & $-$1.4 & $-$1.1 & $-$0.7 & $-$0.7 & $-$1.0 & $-$1.1 & $-$0.3 \\
			Zipf slope & $-$0.6 & $-$0.3 & $-$0.3 & $-$0.8 & $-$0.5 & $-$0.3 & $-$0.5 & $-$0.5 & \textbf{1.8} & \textbf{2.5} \\
			Word length & 1.0 & 0.5 & 0.0 & 0.1 & 0.0 & 0.6 & 0.8 & 0.6 & $-$0.8 & $-$0.9 \\
			Spread of word length & $-$0.4 & $-$1.1 & $-$0.6 & $-$0.9 & $-$0.4 & $-$0.7 & $-$0.8 & $-$0.8 & 0.4 & 0.2 \\
			Length and frequency & 0.1 & $-$0.4 & $-$0.2 & 0.1 & 0.2 & 0.2 & $-$0.3 & 0.4 & 0.4 & 0.3 \\
			Adjacent-word information & $-$0.5 & $-$0.2 & 1.8 & 2.2 & 0.4 & $-$0.1 & 1.2 & 0.9 & 0.9 & \textbf{1.9} \\
			Glyph entropy & $-$0.1 & $-$0.1 & $-$0.1 & $-$0.6 & 0.2 & $-$0.5 & 0.2 & $-$0.5 & $-$1.5 & \textbf{$-$1.7} \\
			Glyph predictability & $-$0.8 & $-$0.7 & $-$1.3 & \textbf{$-$2.3} & $-$1.1 & \textbf{$-$2.0} & $-$1.5 & $-$1.9 & $-$1.1 & \textbf{$-$1.6} \\
			Glyph position rigidity & 0.4 & 0.3 & 0.5 & 1.1 & 0.2 & 1.5 & 1.0 & 1.3 & 0.6 & 0.2 \\
			\midrule
			\multicolumn{11}{@{}l}{\textit{Held-out statistics}} \\
			Vocabulary growth & \textbf{2.5} & \textbf{2.2} & 0.7 & 0.1 & 0.8 & 0.9 & 0.5 & 0.1 & \textbf{3.6} & \textbf{3.3} \\
			Types that occur twice & \textbf{$-$1.9} & $-$1.7 & $-$1.0 & $-$0.9 & $-$1.5 & $-$1.6 & $-$1.0 & $-$1.6 & \textbf{$-$5.5} & \textbf{$-$5.7} \\
			Top 100 types & $-$0.1 & $-$0.5 & $-$0.3 & 0.1 & $-$0.1 & 0.0 & $-$0.1 & 0.0 & $-$0.8 & $-$0.7 \\
			Distance back to a repeat & 0.0 & 0.7 & 1.2 & 1.2 & 0.6 & 0.3 & 0.5 & 0.8 & $-$1.2 & $-$1.4 \\
			Near neighbour among last 10 & 1.2 & 1.4 & $-$0.7 & $-$0.5 & $-$0.4 & 0.0 & $-$0.8 & $-$0.1 & 0.8 & 1.2 \\
			Repeat within a line & 0.8 & $-$0.4 & $-$1.0 & $-$1.0 & $-$0.5 & $-$0.3 & $-$0.6 & $-$0.7 & 2.1 & \textbf{2.6} \\
			Reused word pairs & $-$0.4 & $-$0.8 & $-$0.5 & $-$0.1 & $-$0.4 & $-$0.2 & $-$0.2 & $-$0.5 & 1.5 & \textbf{2.5} \\
			Word entropy given previous & $-$0.6 & $-$1.0 & $-$0.6 & 0.1 & $-$0.2 & $-$0.1 & $-$0.5 & 0.0 & $-$0.5 & 0.0 \\
			Overlap of adjacent pages & 1.6 & 1.1 & 0.5 & 0.3 & \textbf{5.4} & \textbf{4.3} & 1.2 & 1.0 & \textbf{10.3} & \textbf{9.9} \\
			Short words & $-$0.6 & $-$0.8 & $-$0.5 & $-$1.0 & $-$0.3 & $-$0.7 & $-$1.1 & $-$1.0 & 1.2 & 1.6 \\
			Long words & 0.9 & 0.3 & $-$0.1 & $-$0.2 & 0.0 & 0.2 & 0.5 & 0.4 & 0.1 & $-$0.6 \\
			Shared contexts of near neighbours & \textbf{8.2} & \textbf{4.8} & 0.2 & 0.2 & 0.3 & 0.7 & 0.4 & 0.2 & \textbf{5.3} & \textbf{4.3} \\
			Vocabulary drift & \textbf{$-$2.2} & \textbf{$-$1.8} & $-$0.9 & $-$0.7 & \textbf{$-$7.5} & \textbf{$-$5.2} & $-$1.6 & $-$1.3 & \textbf{$-$15.3} & \textbf{$-$14.6} \\
			Currier A/B word divergence & 1.2 & 0.7 & $-$0.1 & $-$0.1 & \textbf{4.7} & \textbf{3.2} & 0.3 & 0.0 & \textbf{12.8} & \textbf{13.6} \\
			Currier A/B glyph divergence & \textbf{3.5} & \textbf{3.8} & $-$0.1 & 0.0 & \textbf{54.5} & \textbf{33.7} & 0.0 & 0.1 & \textbf{196} & \textbf{186} \\
			\midrule
			\multicolumn{11}{@{}l}{\textit{Layout statistics (reported only)}} \\
			Line-initial glyphs & \textbf{343} & \textbf{397} & \textbf{321} & \textbf{358} & \textbf{364} & \textbf{322} & \textbf{300} & \textbf{401} & \textbf{425} & \textbf{363} \\
			Line-final glyphs & \textbf{141} & \textbf{145} & \textbf{122} & \textbf{129} & \textbf{163} & \textbf{133} & \textbf{133} & \textbf{140} & \textbf{196} & \textbf{167} \\
			Paragraph-initial glyphs & \textbf{161} & \textbf{179} & \textbf{144} & \textbf{162} & \textbf{142} & \textbf{149} & \textbf{155} & \textbf{179} & \textbf{169} & \textbf{188} \\
			Line-initial length & \textbf{16.3} & \textbf{17.2} & \textbf{16.4} & \textbf{17.1} & \textbf{15.4} & \textbf{15.3} & \textbf{17.9} & \textbf{16.9} & \textbf{16.1} & \textbf{19.3} \\
			\bottomrule
		\end{tabular*}

	\end{table*}
	
	\appsection{Corpora}
	\label{app:corpora}
	
	Table~\ref{tab:corpora} lists the texts of every language. Documents that train the language models never contribute to the control texts. The letter test of Section~\ref{sec:reading} draws its own split of the same collections, with 100,000 to 1.5 million words of training text per language. For Italian and Middle French it uses instead 6 and 16 documents of edited texts and early prints of works written between 1300 and 1500, all from Project Gutenberg. On them the solver reads 77\% and 84\% of the solved tokens of the controls correctly, against 47\% and 62\% on the machine transcriptions, and the gaps of the controls rise from 0.28 and 0.27 to 0.59 and 0.64. Table~\ref{tab:sources} gives the addresses of the public collections.
	\begin{table*}[t]
		\centering
		\scriptsize
		\setlength{\tabcolsep}{3pt}
				\caption{Texts of the 16 languages and of abbreviated Latin. The main corpus holds texts written between 1300 and 1500. One Middle English text comes from Project Gutenberg and one Middle French text from EEBO-TCP. Hebrew, Arabic and Persian are written as consonants, and Greek and Slavonic are transliterated. The larger models of abbreviated Latin in Section~\ref{sec:wordcodes} use a further 5 million words from 144 Latin manuscripts of CoMMA written between 1200 and 1500.}
		\label{tab:corpora}
		\begin{tabular}{@{}lllrrrr@{}}
			\toprule
			& & & \multicolumn{2}{c}{Letter and language models} & \multicolumn{2}{c}{Control texts} \\
			\cmidrule(lr){4-5}\cmidrule(lr){6-7}
			Language & Source & Dates & documents & letters & documents & words \\
			\midrule
			Arabic & OpenITI & 1022--1340 & 9 & 1,739,547 & 3 & 113,706 \\
			Church Slavonic & PROIEL, TOROT & 1000--1600 & 21 & 1,682,642 & 7 & 100,213 \\
			German & main corpus & 1300--1500 & 4 & 284,379 & 4 & 86,091 \\
			Middle English & main corpus & 1300--1500 & 162 & 3,512,777 & 61 & 80,068 \\
			Persian & Ganjoor & 11th--14th c. & 5 & 1,353,562 & 4 & 82,730 \\
			Middle French & main corpus & 1300--1500 & 10 & 2,963,157 & 3 & 489,819 \\
			Old French & UD PROFITEROLE & 1050--1300 & 6 & 419,694 & 3 & 96,495 \\
			Greek & First1KGreek, PROIEL & 1000--1477 & 5 & 324,138 & 3 & 495,796 \\
			Hebrew & Sefaria & 1075--1550 & 7 & 1,517,760 & 3 & 164,805 \\
			Italian & main corpus & 1300--1500 & 6 & 3,779,487 & 3 & 82,157 \\
			Latin & main corpus & 1300--1500 & 91 & 3,293,556 & 14 & 84,732 \\
			Latin, abbreviated & CREMMA, HTRomance, Paris Bible & medieval & 15 & 181,294 & 48 & 127,745 \\
			Old Norse & Menotec, Heimskringla & 1200--1275 & 25 & 799,786 & 5 & 91,562 \\
			Czech & Kralice Bible & 1579--1613 & 56 & 2,343,350 & 8 & 92,841 \\
			Old Spanish & ISWOC, Project Gutenberg & 1200--1554 & 6 & 692,521 & 4 & 129,197 \\
			Portuguese & ISWOC, Project Gutenberg & 1344--1520 & 4 & 748,116 & 3 & 90,807 \\
			Old Occitan & UD CorAG, TrovaCor & 1100--1564 & 179 & 1,701,048 & 49 & 84,669 \\
			\bottomrule
		\end{tabular}

	\end{table*}
	\begin{table*}[t]
		\centering
		\scriptsize
		\setlength{\tabcolsep}{3pt}
				\caption{Public collections of the texts and the scans of the manuscript.}
		\label{tab:sources}
		\begin{tabular}{@{}ll@{}}
			\toprule
			Collection & Address \\
			\midrule
			Sefaria & \href{https://github.com/Sefaria/Sefaria-Export-Archive}{\nolinkurl{github.com/Sefaria/Sefaria-Export-Archive}} \\
			OpenITI & \href{https://github.com/OpenITI}{\nolinkurl{github.com/OpenITI}} \\
			Ganjoor & \href{https://github.com/ganjoor/ganjoor-db}{\nolinkurl{github.com/ganjoor/ganjoor-db}} \\
			First1KGreek & \href{https://github.com/OpenGreekAndLatin/First1KGreek}{\nolinkurl{github.com/OpenGreekAndLatin/First1KGreek}} \\
			PROIEL, TOROT, ISWOC, Menotec & \href{https://github.com/syntacticus/syntacticus-treebank-data}{\nolinkurl{github.com/syntacticus/syntacticus-treebank-data}} \\
			UD PROFITEROLE & \href{https://github.com/UniversalDependencies/UD_Old_French-PROFITEROLE}{\nolinkurl{github.com/UniversalDependencies/UD_Old_French-PROFITEROLE}} \\
			UD CorAG & \href{https://github.com/UniversalDependencies/UD_Old_Occitan-CorAG}{\nolinkurl{github.com/UniversalDependencies/UD_Old_Occitan-CorAG}} \\
			TrovaCor & \href{https://github.com/Loicdelanghe/TrovaCor}{\nolinkurl{github.com/Loicdelanghe/TrovaCor}} \\
			Project Gutenberg & \href{https://github.com/GITenberg}{\nolinkurl{github.com/GITenberg}} \\
			Heimskringla (CLTK) & \href{https://github.com/cltk/old_norse_texts_heimskringla}{\nolinkurl{github.com/cltk/old_norse_texts_heimskringla}} \\
			Kralice Bible & \href{https://github.com/seven1m/open-bibles}{\nolinkurl{github.com/seven1m/open-bibles}} \\
			CoMMA & \href{https://huggingface.co/datasets/comma-project/comma-jsonl}{\nolinkurl{huggingface.co/datasets/comma-project/comma-jsonl}} \\
			Baumbank.UP & \href{https://hdl.handle.net/11022/0000-0007-EAF7-B}{\nolinkurl{hdl.handle.net/11022/0000-0007-EAF7-B}} \\
			CME & \href{https://quod.lib.umich.edu/c/cme}{\nolinkurl{quod.lib.umich.edu/c/cme}} \\
			PCEEC & \href{https://github.com/beatrice57/pceec2}{\nolinkurl{github.com/beatrice57/pceec2}} \\
			Helsinki Corpus & \href{https://varieng.helsinki.fi/CoRD/corpora/HelsinkiCorpus/}{\nolinkurl{varieng.helsinki.fi/CoRD/corpora/HelsinkiCorpus}} \\
			EEBO-TCP & \href{https://textcreationpartnership.org}{\nolinkurl{textcreationpartnership.org}} \\
						Gibberish of Gaskell and Bowern & \href{https://github.com/danielgaskell/voynich}{\nolinkurl{github.com/danielgaskell/voynich}} \\
			Scans of Beinecke MS 408 & \href{https://collections.library.yale.edu/catalog/2002046}{\nolinkurl{collections.library.yale.edu/catalog/2002046}} \\
			\bottomrule
		\end{tabular}

	\end{table*}

	\appsection{Word Pairs by Language}
	\label{app:pairs}
	
	Table~\ref{tab:pairslang} gives the word-pair replication of the plaintexts of every language and of their word codes. Averaged over both programs and fits, the word codes of each language keep between 58 and 86 percent of the replication of their plaintext at the fitted rate. Only abbreviated Latin and Old Occitan come near ZL before heavy miscopying. With pairs at least two words apart, ZL reaches 0.028 and the natural texts 0.055 to 0.325. With the 150 or 600 most frequent types, ZL reaches 0.069 and 0.070 and the natural texts at least 0.138. The overdispersion of ZL in lines is 0.731, against 0.895 to 2.133 in the natural texts. On pages the replication of ZL is 0.021 to 0.024, depending on the seed of the shuffles, near the lowest value among the natural texts (0.021).
	
	\begin{table}[t]
		\centering
		\scriptsize
		\setlength{\tabcolsep}{2.5pt}
				\caption{Word-pair replication of the enciphered plaintexts and of their word codes (Cipher-W and Cipher-W2, both fits), by miscopying. ZL reaches 0.071.}
		\label{tab:pairslang}
		\begin{tabular}{@{}lcccccc@{}}
			\toprule
			& & \multicolumn{5}{c}{Word codes, miscopied words} \\
			\cmidrule(l){3-7}
			Language & Plain & 0\% & fitted & 20\% & 40\% & 60\% \\
			\midrule
			Arabic & 0.334 & 0.307 & 0.282 & 0.238 & 0.177 & 0.070 \\
			Church Slavonic & 0.178 & 0.135 & 0.144 & 0.102 & 0.044 & 0.026 \\
			German & 0.256 & 0.187 & 0.177 & 0.127 & 0.066 & 0.026 \\
			Middle English & 0.380 & 0.296 & 0.279 & 0.218 & 0.136 & 0.064 \\
			Persian & 0.232 & 0.168 & 0.156 & 0.127 & 0.064 & 0.027 \\
			Middle French & 0.205 & 0.148 & 0.134 & 0.108 & 0.060 & 0.034 \\
			Old French & 0.292 & 0.203 & 0.192 & 0.144 & 0.092 & 0.037 \\
			Greek & 0.272 & 0.191 & 0.194 & 0.137 & 0.085 & 0.050 \\
			Hebrew & 0.218 & 0.200 & 0.188 & 0.155 & 0.081 & 0.036 \\
			Italian & 0.217 & 0.188 & 0.161 & 0.107 & 0.049 & 0.035 \\
			Latin & 0.251 & 0.191 & 0.172 & 0.120 & 0.068 & 0.029 \\
			Latin, abbreviated & 0.102 & 0.078 & 0.074 & 0.076 & 0.029 & 0.029 \\
			Old Norse & 0.252 & 0.178 & 0.181 & 0.130 & 0.078 & 0.039 \\
			Czech & 0.263 & 0.179 & 0.154 & 0.109 & 0.037 & 0.032 \\
			Old Spanish & 0.229 & 0.154 & 0.136 & 0.103 & 0.067 & 0.029 \\
			Portuguese & 0.312 & 0.225 & 0.203 & 0.168 & 0.083 & 0.045 \\
			Old Occitan & 0.177 & 0.123 & 0.112 & 0.074 & 0.036 & 0.021 \\
			\bottomrule
		\end{tabular}

	\end{table}
		\begin{table}[t]
		\centering
		\small
				\caption{Sequences of three words that come back ten or more pages later, with one word allowed to differ, per 1,000 sequences and above the rate in the page-shuffled text. The plaintexts give their range and the programs the central 95\% of their texts. HABIT and SPACE were fitted without the six measures of coupling, and the nomenclators miscopy at their fitted rates. Bold marks ranges that exclude ZL.}
		\label{tab:returns}
		\begin{tabular}{@{}lr@{}}
			\toprule
			& Returning sequences \\
			Texts & (per 1,000) \\
			\midrule
			ZL & 23.5 \\
			IT & 24.2 \\
			\midrule
			29 plaintexts & $-$20.1--79.6 \\
			\midrule
			Copy-T & \textbf{$-$37.4--2.8} \\
			Copy-H & \textbf{7.4--15.7} \\
			HABIT & $-$13.0--30.7 \\
			SPACE & \textbf{$-$30.8--10.3} \\
			\midrule
			Cipher-M & $-$8.6--32.3 \\
			Cipher-M2 & 4.4--29.9 \\
			Cipher-MN & 10.1--37.7 \\
			\bottomrule
		\end{tabular}

	\end{table}
	\appsection{The Open Nomenclator by Language}
	\label{app:nomen}
	Table~\ref{tab:nomen} gives the decisions of Section~\ref{sec:nomen} for every language. For each parameter set it gives a margin, the lowest gap of its two controls in three readings minus the highest gap of ZL. The margin is taken under the solver that comes closest to excluding the set, and a positive margin excludes the set. For the third set the table also gives the lowest share of the most frequent word among its 100 texts minus that of ZL, in Currier A and in Currier B.
	\begin{table}[t]
		\centering
		\scriptsize
		\setlength{\tabcolsep}{2.5pt}
				\caption{The second fit of Cipher-M2 by language, for ZL. The margins of the solvers are in the units of the gap, and those of the most frequent word are shares of the words of Currier A / Currier B. The decisions for IT are the same.}
		\label{tab:nomen}
		\begin{tabular}{@{}lrrrcl@{}}
			\toprule
			& \multicolumn{3}{c}{Margin under the solvers} & Most frequent & \\
			\cmidrule(lr){2-4}
			Language & Set 1 & Set 2 & Set 3 & word, set 3 & Cipher \\
			\midrule
						Arabic & $-$0.041 & 0.132 & $-$0.081 & 0.049 / 0.065 & not excluded \\
			Church Slavonic & 0.122 & 0.451 & $-$0.027 & 0.049 / 0.067 & excluded \\
			German & 0.115 & 0.266 & $-$0.058 & 0.045 / 0.063 & excluded \\
			Middle English & 0.057 & 0.316 & $-$0.101 & 0.045 / 0.064 & excluded \\
			Persian & 0.094 & 0.155 & $-$0.056 & 0.048 / 0.064 & excluded \\
			Middle French & 0.033 & 0.347 & $-$0.046 & 0.052 / 0.067 & excluded \\
			Old French & 0.157 & 0.259 & $-$0.090 & 0.045 / 0.066 & excluded \\
			Greek & 0.270 & 0.703 & 0.083 & 0.053 / 0.065 & excluded \\
			Hebrew & $-$0.107 & $-$0.110 & $-$0.109 & 0.047 / 0.064 & undecided \\
			Italian & 0.087 & 0.318 & $-$0.004 & 0.051 / 0.062 & excluded \\
			Latin & 0.040 & 0.301 & $-$0.039 & 0.046 / 0.063 & excluded \\
			Old Norse & 0.106 & 0.327 & $-$0.107 & 0.045 / 0.065 & excluded \\
			Czech & 0.244 & 0.481 & 0.048 & 0.047 / 0.066 & excluded \\
			Old Spanish & 0.467 & 0.792 & 0.118 & 0.048 / 0.062 & excluded \\
			Portuguese & 0.619 & 0.968 & 0.254 & 0.048 / 0.064 & excluded \\
			Old Occitan & 0.217 & 0.447 & $-$0.049 & 0.048 / 0.065 & excluded \\
			\bottomrule
		\end{tabular}

	\end{table}
	
	\appsection{Implementation}
	\label{app:impl}
	All programs, statistics and solvers are written in Python with NumPy, and we will release the code and decision rules with the published version. The experiments ran on a queue of tasks shared by several computers, and every simulation is cached under a hash of its program, parameters and seed. Every decision rule in this paper was fixed in the code before the experiment that applies it was run, except the combined rule for the open nomenclator in Section~\ref{sec:nomen}.
	
	\paragraph{Earlier attempts}
	Before the experiments reported here, we tried two other routes to a reading. The first learned rules that rewrite manuscript words into words of up to nine languages, from word embeddings aligned across languages, choosing rule tables by integer programming. It recovered substitution ciphers of known texts. On the manuscript no rule table met the criteria we had fixed, and the tables learned from ZL and from IT often turned the same word into different words. Six of the seven word meanings it proposed were also given to many other manuscript words by the same table. The second route assigned meanings to manuscript words with small language models trained on historical texts and Bibles and with XLM-R \citep{conneau2020xlmr}. On held-out words of known languages it found the labelled meaning about as often as a match by spelling alone, and it did not reliably recover a known text enciphered word by word. Pages of the manuscript guided the development of both routes, so we treat their results as exploratory. Two later tests ran in five languages only. A solver that reads each frequent word as a syllable read between 9 and 81 percent of the syllables of ten Cipher-S controls, and the gap of the manuscript was smaller than that of every control, even of controls read with the model of another language. An earlier version of the filler test had seven other rules, which drop repeated or predictable words, merge runs and spelling variants or delete predictable glyphs. None raised the word-pair replication of ZL by more than 0.021, less than the rule that removes near copies in Section~\ref{sec:other}.
	
	\begin{acknowledgments}
	We thank Igor E. Golub (independent researcher, France) for reproducing the word-pair replication independently and for noticing that our first description of the near-copies we leave out did not match the code. We used Claude (Anthropic) to check and edit the code of the experiments and of the figures and tables, to suggest some of the experiments and to polish the wording of the text. The authors take full responsibility for the content of the paper.
	\end{acknowledgments}
	
	\bibliographystyle{compling}
\bibliography{COLI_template}

\begin{thebibliography}{77}
\bibitem[{Rozanova and Temerev(2026)}]{rozanova2026glyph}
Rozanova, Liudmila and Alexander Temerev. 2026.
\newblock A glyph is not a letter, a token is not a word, a space is not a
  space: What the units of {V}oynichese are not.
\newblock arXiv preprint arXiv:2608.17096.
\end{thebibliography}
\end{document}